\pdfoutput=1
\documentclass[11pt]{article}

\usepackage[dvipsnames, table]{xcolor}
\usepackage{bm}
\usepackage[final]{naacl2021}
\usepackage{amsfonts}
\usepackage{bbm}
\usepackage{graphicx}
\usepackage{amsmath}
\usepackage{times}
\usepackage{tcolorbox}
\tcbuselibrary{breakable, skins}
\usepackage{latexsym}
\usepackage{booktabs}
\usepackage{multirow}
\usepackage[T1]{fontenc}
\usepackage[utf8]{inputenc}
\usepackage{microtype}
\usepackage{array}
\usepackage{booktabs}
\usepackage{colortbl}
\usepackage{subcaption}
\usepackage{tabularray}
\usepackage{placeins}
\usepackage{float}

\newtcolorbox{promptbox}[1]{
    breakable,
    colback=gray!5,
    colframe=gray!50!black,
    fonttitle=\bfseries,
    title=#1,
    attach boxed title to top left={yshift=-3mm, xshift=2mm},
    boxed title style={size=small, colback=gray!50!black}
}

\definecolor{lightred}{RGB}{240, 180, 180}
\definecolor{lightgreen}{RGB}{150, 243, 150}

\newcommand{\colorcellgreen}[3]{%
  \pgfmathsetmacro{\perc}{(1-(#1-#2)/(#3-#2))*100}%
  \edef\cellcolorcmd{\noexpand\cellcolor{white!\perc!lightgreen}}%
  \cellcolorcmd
}

\newcommand{\colorcellred}[3]{%
  \pgfmathsetmacro{\perc}{((#1-#2)/(#3-#2))*100}%
  \edef\cellcolorcmd{\noexpand\cellcolor{white!\perc!lightred}}%
  \cellcolorcmd
}

\title{Tokenization Constraints in LLMs:\\ A Study of Symbolic and Arithmetic Reasoning Limits}

\author{
Xiang Zhang$^{1}$\thanks{$^{*}$Equal contribution.} \qquad Juntai Cao$^{1*}$\qquad Jiaqi Wei$^2$ \qquad Yiwei Xu$^{3}$ \qquad Chenyu You$^{4}$ \\ 
$^{1}$ University of British Columbia \\
$^{2}$ Zhejiang University \\
$^{3}$ Cisco\\
$^{4}$ Stony Brook University \\ 
\texttt{xzhang23@ualberta.ca, jtcao7@cs.ubc.ca, chenyu.you@stonybrook.edu}}

\begin{document}

\maketitle
\begin{abstract}
Tokenization is the first—and often underappreciated—layer of computation in language models. While Chain-of-Thought (CoT) prompting enables transformer models to approximate recurrent computation by externalizing intermediate steps, we show that the success of such reasoning is fundamentally bounded by the structure of tokenized inputs. This work presents a theoretical and empirical investigation into how tokenization schemes, particularly subword-based methods like byte-pair encoding (BPE), impede symbolic computation by merging or obscuring atomic reasoning units. We introduce the notion of \textit{Token Awareness} to formalize how poor token granularity disrupts logical alignment and prevents models from generalizing symbolic procedures. Through systematic evaluation on arithmetic and symbolic tasks, we demonstrate that token structure dramatically affect reasoning performance, causing failure even with CoT, while atomically-aligned formats unlock strong generalization, allowing small models (e.g., GPT-4o-mini) to outperform larger systems (e.g., o1) in structured reasoning. Our findings reveal that symbolic reasoning ability in LLMs is not purely architectural, but deeply conditioned on token-level representations. Full code, prompts, and results are available at \href{https://anonymous.4open.science/r/Counting-ARR-DDEF}{Anonymous GitHub}.
\end{abstract}

\section{Introduction}

\textit{Inductive reasoning and arithmetic computation}, such as counting, addition, and pattern generalization, are foundational components of symbolic and algorithmic intelligence. These abilities have long been studied across disciplines—from their cognitive development in humans~\cite{wynn1990children,de1964polya} to their formal characterization in logic and computability theory~\cite{boolos2002computability,cooper2017computability}. In theoretical computer science, arithmetic primitives like counting and addition have been analyzed in terms of circuit complexity~\cite{jerrum1995computational}, computational depth~\cite{fischer1968counter}, and machine models such as counter automata~\cite{ibarra2002counter}. Even simple operations—e.g., counting from \texttt{1} to \texttt{n}—are known to require a depth complexity that grows with input length, imposing strict lower bounds on any computational model~\cite{fischer1968counter}.

Transformers~\cite{vaswani2017attention}, including both autoregressive (GPT-style) \cite{gregor2014deep,achiam2023gpt} and non-autoregressive (BERT-style) \cite{devlin2018bert,liu2022character} variants, are inherently limited to \textit{constant-depth} computation~\cite{zhang2024autoregressivechainthought,deletang2022neural,li2024chain}. In neural models, all computation occur within model's latent space $\bm{\mathcal{H}}$, where the hidden state $\bm{h}$ encodes intermediate computational representations. Unlike recurrent architectures—where hidden states evolve over time via recursive updates $\bm{h}_t = g_\theta(\bm{h}_{t-1})$—Transformers update $\bm{h}$ only across a fixed number of layers, independent of input length. As a result, a standard Transformer can process (or reason over) its hidden states only a constant number of times, limiting its computational depth and situating it at the lower end of the Chomsky hierarchy~\cite{deletang2022neural}. This architectural bottleneck fundamentally restricts Transformer-based models—ranging from task-specific expert systems to large-scale LLMs—from solving even basic arithmetic operations such as \textit{counting}, which require iterative updates to internal state and growing depth with input length.

\begin{figure*}[t!]
    \centering
    \includegraphics[width=0.8\linewidth]{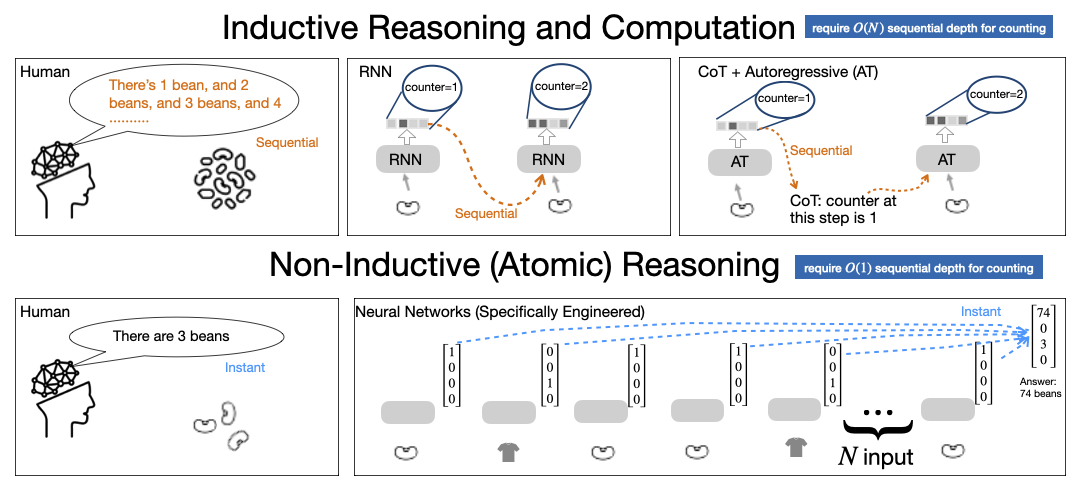}
    \caption{Illustration of inductive reasoning as performed by humans, RNNs, and LLMs with CoT, respectively. }
    \label{fig:inductive}
    \vspace{-1.em}
\end{figure*}
Chain of Thought (CoT)\cite{wei2022chain} revolutionizes the reasoning paradigm by shifting the locus of computation from the latent space $\bm{\mathcal{H}}$ to the textual output space $\bm{\mathcal{O}}$\cite{zhang2024autoregressivechainthought}. By externalizing intermediate reasoning steps into natural language "thoughts," CoT enables transformer-based models to tackle fundamental computational tasks that would otherwise exceed their architectural capacity. These include basic arithmetic and symbolic reasoning operations such as counting, addition, and sequence manipulation. Theoretical studies~\cite{li2024chain,zhang2024autoregressivechainthought,feng2024towards} demonstrate that CoT-augmented language models, under idealized assumptions, possess an \textit{upper bound} capacity to simulate computations of arbitrary complexity—thereby extending the class of tasks solvable beyond what standard Transformers can achieve.

Despite extensive theoretical analyses and guarantees on the upper bound of computational abilities~\cite{zhang2024autoregressivechainthought,chang2024language}, actual model performance remains far below these limits. As LLMs scale from millions to billions of parameters~\cite{achiam2023gpt}, improvements on fundamental tasks such as counting have been marginal—GPT-4, for instance, still struggles to count the number of "r"s in a word. While recent work has explored contributing factors like training data~\cite{allen2023physics} and positional encoding~\cite{chang2024language}, one of the most basic components—tokenization—has received surprisingly little attention. In particular, modern byte pair encoding (BPE)~\cite{sennrich2015neural} merges multiple characters into single tokens for efficiency, often degrading arithmetic reasoning due to information loss during tokenization. Even OpenAI’s latest o1\footnote{\url{https://openai.com/o1}} model, which integrates Monte Carlo Tree Search (MCTS) for improved reasoning, achieves only 50\% accuracy on long string arithmetic tasks involving 30–40 characters.

In this work, we systematically investigate how tokenization choices can substantially constrain the theoretical reasoning and arithmetic capabilities of neural models. Our approach is model-agnostic, allowing us to evaluate even closed-source LLMs with undisclosed tokenization schemes. Leveraging extensive experiments with Chain of Thought (CoT)—which has been theoretically shown to achieve Turing completeness under idealized conditions~\cite{li2024chainthoughtempowerstransformers}—we demonstrate that the choice of tokenization plays a critical role in unlocking a model’s full computational potential and bridging the gap between theoretical guarantees and practical performance. Neglecting this factor can lead to performance degradations of up to 80\%. Moreover, we find that the impact of tokenization is model-dependent: some tokens disproportionately hinder performance on counting tasks, even when the underlying task remains fixed.


\section{Neural Networks and Arithmetic: A Revisit}
\label{sec:relatedwork}

\textbf{Training neural networks for arithmetic computation.} Arithmetic operations—including counting, matching, and bracket balancing—are foundational for symbolic reasoning and more complex algorithmic tasks~\cite{chang2024language}. Early studies on training neural networks (NNs) for such tasks focused on architectures capable of handling variable-length inputs. Since multi-layer perceptrons (MLPs)~\cite{rosenblatt1958perceptron} are inherently limited to fixed-size inputs, initial progress came through recurrent neural networks (RNNs).

\citet{rodriguez1999recurrent} trained early RNNs to recognize the \textit{regular} language $\texttt{a}^n\texttt{b}^n$, which requires the network to implicitly count occurrences of \texttt{a} and \texttt{b}. Of the 50 networks trained, 8 successfully generalized to longer sequences, highlighting RNNs' capacity for basic arithmetic generalization. Building on this, \citet{suzgun2019lstmnetworksperformdynamic} showed that LSTMs could perform more complex \textit{dynamic counting} via bracket pairing tasks, leveraging gating and cell-state mechanisms to maintain multiple counters—capabilities that standard RNNs lacked.
\begin{figure}[t]
    \centering
    \includegraphics[width=\linewidth]{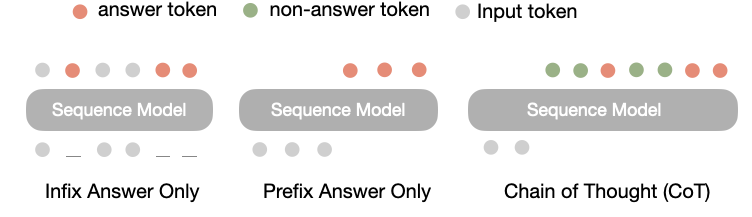}
    \vspace{-0.1em}
    \caption{CoT vs Answer Only Generation Models.}
    \label{fig:anseronly}
    \vspace{-1em}
\end{figure}

\textbf{Theory of Model Architectures with Composability. } \citet{deletang2022neural} systematically investigated arithmetic capabilities across modern architectures, including RNNs, LSTMs, and Transformers. Their findings confirmed that while LSTMs exhibit computational behavior aligned with counter machines, Transformers consistently fail at even basic counting tasks. \citet{chang2024language} extended this analysis to newer architectures such as Mamba~\cite{gu2023mamba} and RWKV~\cite{peng2023rwkv}, revealing that these models also underperform on arithmetic tasks outside their training distributions—often performing worse than classic RNNs in generalization.

Recent studies~\cite{weiss2018practicalcomputationalpowerfinite,ackerman2020survey} have further validated the computational capabilities of both RNNs and LSTMs, particularly for tasks requiring symbolic or arithmetic reasoning. In contrast, Transformers—lacking inherent recurrence—are restricted to TC$^{0}$ complexity in their inductive reasoning capacity~\cite{li2024chain}, placing them at the lower bound of the Chomsky hierarchy~\cite{sanford2024transformers,li2024chain,deletang2022neural}. As a result, they are fundamentally incapable of solving even basic algorithmic tasks, such as arithmetic pattern induction or sequence manipulation, without incorporating explicit inductive biases~\cite{chang2024language}.

\section{Theoretical Limits of Answer-Only Models for Arithmetic and Symbolic Computation.}

Transformer-based models~\cite{vaswani2017attention} without Chain-of-Thought (CoT) prompting (Fig.~\ref{fig:anseronly} left 2) are inherently constrained by their \textit{fixed architectural depth} and lack of \textit{recurrence}. Let $\bm{\mathcal{X}}$ be the input token space, $\bm{\mathcal{H}}$ the hidden state space, and $\bm{\mathcal{O}}$ the output token space. For a Transformer with $L$ layers and input sequence $\bm{x}_{1:n} \in \bm{\mathcal{X}}^n$, the hidden representation at layer $\ell$ and position $t$ is given by $\bm{h}_t^{(\ell)} = \text{Layer}_\ell\left(\bm{h}_{1:n}^{(\ell-1)}\right)$, with $\bm{h}_t^{(0)} = \text{Embed}(\bm{x}_t)$. The output token $\bm{o}_t \in \bm{\mathcal{O}}$ is then computed as $\bm{o}_t = \text{Softmax}\left(W\bm{h}_t^{(L)}\right)$.

This process applies a fixed sequence of transformations to each input $\bm{x}_t$, with the number of computational steps bounded above by $L = \mathcal{O}(1)$. Since $L$ does not scale with the input length $n$, the \textit{depth complexity} of computation in such models is constant, i.e., $\text{Depth}_{\text{Transformer}} = \mathcal{O}(1)$. As established in complexity theory~\cite{zhang2024autoregressivechainthought,li2024chain,chang2024language}, this places answer-only Transformers in the class $\text{TC}^0$—constant-depth circuits with polynomial size and threshold gates—incapable of performing even simple arithmetic functions such as parity, addition, or comparison over unbounded inputs.

Formally, consider a function $f: \bm{\mathcal{X}}^n \to \bm{\mathcal{O}}^m$ defined by a task such as computing $\texttt{sum}(\bm{x}_{1:n})$. Such tasks require a computation of depth $\Omega(\log n)$ for associative operations and $\Omega(n)$ for sequentially dependent operations (e.g., counting, carry propagation, or string reversal)~\cite{fischer1968counter}. Since $L$ is constant in Transformers and all transformations are composed in parallel across tokens, such models fail to meet the depth requirement: $\text{Depth}_{\text{task}}(f) > \text{Depth}_{\text{Transformer}}$ implies that the Transformer cannot compute $f$.

Furthermore, Transformers lack a mechanism to store and evolve intermediate computational states over time. In recurrent models, hidden states $\bm{h}_t$ are recursively defined as $\bm{h}_t = g_\theta(\bm{h}_{t-1}, \bm{x}_t)$, allowing the system to simulate Turing-complete behavior~\cite{zhang2024autoregressivechainthought,li2024chainthoughtempowerstransformers}. In contrast, Transformers treat all inputs simultaneously through attention-based aggregation without iterative update: $\bm{h}_t^{(L)} = f_{\text{attn}}(\bm{x}_{1:n})$, disallowing symbolic loop constructs or dynamic memory—key components in arithmetic computation.

When constrained to generate only answer tokens $\bm{o}_{1:m} \in \bm{\mathcal{O}}^m$ without emitting intermediate reasoning steps (Fig~\ref{fig:anseronly}), the model’s total computational budget is tightly bound by $m$, which is typically small. Let $\bm{o}_{1:m} = f(\bm{x}_{1:n})$ be the model’s prediction. Then, for a computation that requires $T(n)$ steps, with $T(n) \gg m$, the model must either (1) compress computation into fixed layers—violating the task's depth complexity—or (2) memorize input-output mappings—an approach that does not generalize beyond training.

This reliance on shallow function approximation implies that such models can only succeed by memorizing specific input-output pairs, not by executing general algorithms. Empirically, these models exhibit sharp performance degradation on arithmetic tasks outside their training distribution~\cite{chang2024language}.

If  the maximum number of distinct computation traces a model can represent with $d$-dimensional hidden states and $p$-bit numerical precision: $|\bm{\mathcal{H}}| \leq 2^{dp}$. Therefore, the number of unique state transitions is also bounded above by $2^{dp}$, which is finite and insufficient for representing the $\mathcal{O}(n)$-length trajectories required for tasks such as binary addition or bracket matching.

In summary, models limited to generating only final answer tokens without CoT or external recurrence simulation (Fig~\ref{fig:anseronly}).

\section{CoT under Ideal Assumptions Enables General Arithmetic Computation}
\label{sec:cot_arithmetic_theory}

Transformer-based LLMs~\cite{achiam2023gpt,touvron2023llama,bai2023qwen}, though powerful, are fundamentally bounded by their \textit{fixed architectural depth}~\cite{li2024chain,zhang2024autoregressivechainthought}, limiting their ability to perform arithmetic operations that require sequential, stateful updates. Chain-of-Thought (CoT) prompting~\cite{wei2022chain}, however, offers a mechanism to simulate recurrence, transforming the depth-limited Transformer into a theoretically Turing-complete system under ideal assumptions.

\subsection{Inductive Arithmetic Requires Depth}
\label{subsec:inductive}

Arithmetic reasoning, in both human cognition and formal computation, often involves \textit{inductive updates} across time or space (Fig~\ref{fig:inductive}). For instance, computing the cumulative sum of a digit sequence $\bm{x}_{1:n} = (\texttt{7}, \texttt{3}, \texttt{2}, \cdots)$ requires maintaining an accumulator that evolves as $\bm{s}_t = \bm{s}_{t-1} + \bm{x}_t$ over $t = 1$ to $n$. In recurrent neural networks (RNNs), this is naturally represented as
$\bm{h}_t = g_\theta(\bm{h}_{t-1}, \bm{x}_t)$, 
where the hidden state $\bm{h}_t$ stores intermediate quantities such as partial sums, carries, or flags. The computation depth required for such tasks is $\mathcal{O}(n)$, aligning with results from counter machine theory~\cite{fischer1968counter}.

Transformers, in contrast, lack temporal recurrence. Their hidden states $\bm{h}_t$ are updated via a fixed sequence of layers, independent of sequence length. As a result, their total reasoning depth is $\mathcal{O}(1)$ per token.
Because all $\bm{x}_t$ are processed in parallel, the Transformer cannot simulate stepwise updates required for arithmetic unless all logic is memorized or encoded through exponentially wide circuits~\cite{li2024chain}.

This explains why arithmetic tasks—such as computing $\texttt{sum}(\bm{x}_{1:n})$, $\texttt{reverse}(\bm{x}_{1:n})$, or $\texttt{count}_{\texttt{token}}(\bm{x}_{1:n})$—are infeasible for answer-only Transformers without inductive bias~\cite{chang2024language,deletang2022neural}. These tasks require \textit{depth-sensitive} computation, where each output depends on a chain of intermediate results not recoverable from input alone.

\subsection{Chain-of-Thought Simulates Recurrent Computation}
\label{subsec:cot_recurrent}

Chain-of-Thought (CoT) reasoning allows a Transformer to externalize its hidden state through intermediate tokens. Instead of directly mapping $\bm{x}_{1:n} \rightarrow \bm{y}$, the model generates a sequence of \textit{thought tokens} $\bm{o}_{1:k}$:
\[
\bm{x}_{1:n} \Rightarrow (\bm{o}_1, \bm{o}_2, \dots, \bm{o}_k) \Rightarrow \bm{y}.
\]
Each $\bm{o}_t$ can encode intermediate computations (e.g., partial sums, loop counters, flags) that are later re-ingested through the embedding layer, reinitializing the next latent state:
$
\bm{h}_{t+1} = f_\theta(\texttt{Embed}(\bm{o}_t), \bm{x}_{t+1}).$ 

This iterative reasoning cycle approximates the recurrence in RNNs:$ 
\bm{h}_{t-1} \Rightarrow \bm{o}_t \Rightarrow \bm{h}_t$,  
where $\bm{o}_t$ encodes sufficient information from $\bm{h}_{t-1}$ to resume and advance computation. Under ideal assumptions—namely unlimited CoT token budget and precise token-to-state fidelity—this externalization loop can simulate unbounded depth, making CoT+autoregressive models \textit{Turing complete}~\cite{zhang2024autoregressivechainthought,li2024chainthoughtempowerstransformers}.

\section{Tokenization as a Barrier to Chain-of-Thought Computation}
\label{sec:tokenizer_cot_damage}

Despite the theoretical promise of Chain-of-Thought (CoT) prompting to approximate Turing-complete computation under ideal assumptions~\cite{zhang2024autoregressivechainthought,li2024chain}, empirical failures persist even in state-of-the-art models such as GPT-4. These failures are particularly evident in arithmetic and symbolic tasks that require precise reasoning over fine-grained units (e.g., digits, letters, or symbols), where large language models often yield incorrect results for inputs of even moderate length (e.g., computing the number of \texttt{r}s in \texttt{Strawberry}). This discrepancy highlights a critical limitation: CoT effectiveness is inherently bounded not just by model architecture, but also by the \textit{expressiveness} of the underlying language, which is in turn shaped by the tokenizer.

We treat tokenization as a black-box preprocessor $\bm{x}_{1:n} \xrightarrow[]{\mathcal{T}} \bm{t}_{1:m}$, where $\bm{x}_{1:n}$ is a raw input string over characters and $\bm{t}_{1:m}$ is the resulting sequence of tokens from vocabulary $\bm{\mathcal{V}}$. Modern tokenizers such as byte-pair encoding (BPE)~\cite{sennrich2015neural} are designed to optimize compression and training efficiency, not fidelity of semantic or syntactic granularity. However, for CoT reasoning to succeed in arithmetic and symbolic computation, the token space $\bm{\mathcal{V}}$ must be able to \textit{express and preserve intermediate state information}.

\subsection{Expressiveness and the Token-to-Thought Mapping}
 We define the expressiveness of a language $\mathcal{L} = (\bm{G}, \bm{\mathcal{V}})$—with grammar $\bm{G}$ and vocabulary $\bm{\mathcal{V}}$—as the number of unique semantically meaningful sequences $\bm{S}_{\mathcal{L}}$ it can generate:
\[
\text{Expressiveness}(\mathcal{L}) := |\bm{S}_{\mathcal{L}}|.
\]
For a CoT process to emulate Turing-complete computation, it must support a recurrent approximation: $\bm{h}_{t-1} \rightarrow (\bm{o}_{1},\dots,\bm{o}_k) \rightarrow \bm{h}_t$, where latent state $\bm{h}_{t-1} \in \bm{\mathcal{H}}$ is decoded into intermediate natural language tokens $\bm{o}_{1:k} \in \bm{\mathcal{V}}^k$, which are then re-embedded and fed back to reconstruct $\bm{h}_t$.

This implies the need for a high-fidelity vector-to-token mapping:
\[
\phi: \bm{\mathcal{H}} \longrightarrow \bm{\mathcal{V}}^* \quad \text{and} \quad \psi: \bm{\mathcal{V}}^* \longrightarrow \bm{\mathcal{H}},
\]
such that the composed transformation $\psi \circ \phi(\bm{h}_{t-1}) \approx \bm{h}_t$ retains sufficient computational state to perform stepwise updates. Tokenization introduces two major obstacles to this cycle, degrading the effective CoT expressiveness.

\subsection{Damage Type I: Information Hiding via Token Granularity}
\label{subsec:damage1}
The first form of damage is semantic obfuscation. Suppose the reasoning task requires operating over atomic units (e.g., characters, digits), but tokenization merges these into opaque multi-character tokens: $\texttt{Strawberry} \rightarrow [\texttt{Straw}, \texttt{berry}]$. Now, let $\bm{t}_i \in \bm{\mathcal{V}}$ denote a token for which the model lacks fine-grained internal features (e.g., how many \texttt{r}'s are present). We define the token awareness function as:
\[
\text{TokenAware}(\bm{t}_i, \texttt{prop}) := \mathbb{I}\left[\texttt{prop} \in \text{Emb}(\bm{t}_i)\right],
\]
where $\texttt{prop}$ denotes a property (e.g., digit count, lexical features), and $\text{Emb}(\bm{t}_i)$ is the token embedding. When $\text{TokenAwareness}(\bm{t}_i, \texttt{prop}) = 0$, reasoning that relies on $\texttt{prop}$ (e.g., “count the number of 3’s”) will fail. Thus, even if the CoT reasoning process is intact, its input signal is corrupted at the encoding layer.

\subsection{Damage Type II: Limited CoT Expressiveness via Token Bottleneck}
\label{subsec:damage2}
The second, more subtle limitation occurs during the CoT process itself. The latent state $\bm{h}_{t-1}$ stores accumulated reasoning. To externalize this into thought tokens $\bm{o}_{1:k}$, we require that:
\[
\forall \bm{h}_{t-1} \in \bm{\mathcal{H}}, \exists \bm{o}_{1:k} \in \bm{\mathcal{V}}^k \text{ such that } \phi(\bm{h}_{t-1}) = \bm{o}_{1:k},
\]
but when $\bm{\mathcal{V}}$ is coarse (e.g., BPE with token merges) or lacks the necessary expressive forms (e.g., missing digits, variable names, or operations), this surjection fails. Let $\bm{S}_{\bm{h}}$ be the space of latent states and $\bm{S}_{\bm{o}}$ be the expressible token sequences. Then CoT fidelity is bounded as:
\[
\text{Fidelity}(\text{CoT}) \leq \frac{|\phi(\bm{S}_{\bm{h}}) \cap \bm{S}_{\bm{o}}|}{|\bm{S}_{\bm{h}}|}.
\]
Low expressiveness $\Rightarrow$ low overlap $\Rightarrow$ critical reasoning steps cannot be externalized.

This mismatch becomes catastrophic in arithmetic tasks where thought tokens must verbalize structured computations (e.g., carrying, intermediate sums). Without expressive enough $\bm{\mathcal{V}}$, many $\bm{h}_{t-1}$ are untranslatable, rendering the CoT step ineffective. The model is thus forced to either truncate reasoning or approximate it via memorized heuristics, both of which degrade generalization.

\subsection{Formal Failure Case: CoT under BPE Tokenizer}
Assume the model is asked to compute a symbolic function $f: \bm{x}_{1:n} \rightarrow \bm{y}$ (e.g., reverse digits). The optimal CoT process proceeds via:
\[
\bm{h}_0 \rightarrow \bm{o}_{1:k_1} \rightarrow \bm{h}_1 \rightarrow \bm{o}_{k_1+1:k_2} \rightarrow \cdots \rightarrow \bm{y},
\]
but if $\bm{o}_{1:k_i} \notin \bm{S}_{\mathcal{L}}$ due to token constraints, then $\bm{h}_{i+1}$ will be misaligned, i.e., $\bm{h}_{i+1} \not\approx g(\bm{h}_i)$. Over time, errors compound, and $f$ becomes uncomputable.

\begin{figure*}[t!]
    \centering
    \includegraphics[width=0.85\linewidth]{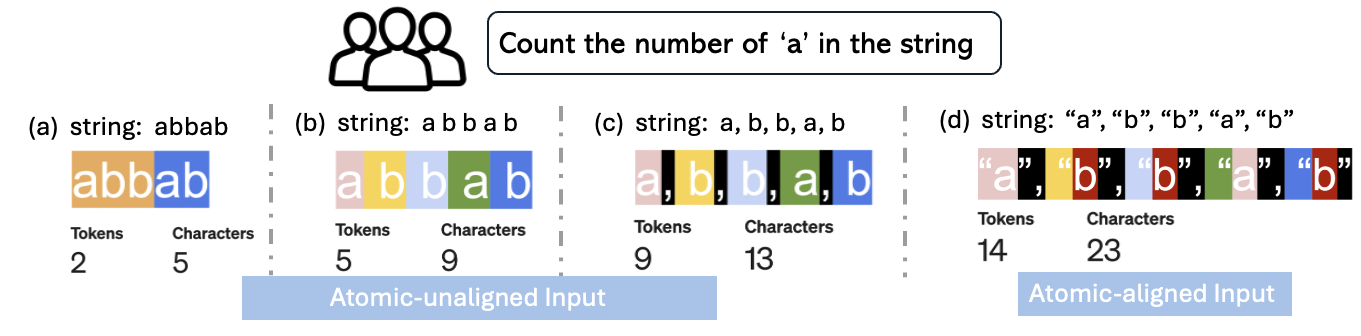}
    \caption{Four types of string formatting to manipulate tokenization in counting. Examples in the figure are tokenized using the GPT-4o tokenizer. Each string-token type is labeled as (a), (b), (c), and (d) in the diagram. Note that changing the format does not alter the fundamental nature or difficulty of the counting task. }
    \label{fig:counting_tokenize}
\end{figure*}

\subsection{Quantifying Tokenization Effects on Symbolic Computation}
\label{sec:quantification_framework}

To complement our theoretical findings, we introduce a general, model-agnostic framework to evaluate how tokenization impacts symbolic and arithmetic reasoning in LLMs. While many models are closed-source, we treat LLMs as black boxes and isolate tokenization as the key variable influencing performance.

Let $\bm{x} \in \bm{\mathcal{X}}^n$ be a character-level input string and $\mathcal{T}$ a tokenizer mapping it into a token sequence $\bm{t}_{1:m} = \mathcal{T}(\bm{x})$, where each $\bm{t}_i \in \bm{\mathcal{V}}$ comes from a fixed vocabulary. The LLM $\mathcal{M}$ then performs a symbolic task by computing an output $\bm{y} = \mathcal{M}(\bm{t}_{1:m})$. Our goal is to determine how mismatches between task granularity and token structure affect the model's ability to solve $f : \bm{\mathcal{X}}^n \rightarrow \bm{\mathcal{Y}}$.

 We base our input manipulation on three typical properties of modern BPE-like tokenizers: (1) Common substrings of 2–4 characters are merged into single tokens. (2) Delimiters (e.g., spaces, commas) are usually merged with adjacent tokens. (3) Adding repeated delimiters can break these merges and force token boundaries.
These patterns let us construct inputs that vary tokenization while keeping the underlying symbolic task fixed.

\paragraph{Input Design.} For a symbolic function $f$ (e.g., digit sum, string reversal, pattern matching), we generate two sets of inputs:
\begin{itemize}
    \item \textbf{Atomic-aligned inputs} $\bm{x}^{\text{atomic}}$: token boundaries align with units required for the task.
    \item \textbf{Merged-token inputs} $\bm{x}^{\text{merged}}$: intentionally merged to obscure symbolic units within tokens.
\end{itemize}

If the model lacks internal awareness of sub-token structure (e.g., characters inside a token), then symbolic reasoning that depends on those units will fail—even if CoT prompting is used.

\paragraph{Quantifying Degradation.} We define the tokenization damage as the average accuracy drop:
\[
\Delta_{\text{tok}} := |\mathbb{E}_{\bm{x}} \left[ \mathcal{A}(\bm{x}^{\text{atomic}}) - \mathcal{A}(\bm{x}^{\text{merged}}) \right]|,
\]
where $\mathcal{A}(\cdot)$ is model accuracy. 

\begin{tcolorbox}[colback=gray!10, colframe=gray!50, boxrule=0.5pt, arc=3pt, left=6pt, right=6pt, top=4pt, bottom=4pt]
\textbf{A high $\Delta_{\text{tok}}$ indicates that the model relies on token structure and cannot generalize symbolic logic across inconsistent tokenizations}.
\end{tcolorbox}

\begin{table*}[t!]
\centering
\resizebox{\textwidth}{!}{
\begin{tabular}{rcccccc|cccccc}
\toprule
 & \multicolumn{6}{c|}{\textbf{Counting letter \texttt{a}}} & \multicolumn{6}{c}{\textbf{Counting letter \texttt{b}}} \\
\cmidrule(lr){2-7} \cmidrule(lr){8-13}
 \multirow{2}{*}{\textbf{String-Token Type \ \ \ }}& \multicolumn{2}{c}{\textbf{len $\in$ [10-20]}} & \multicolumn{2}{c}{\textbf{len $\in$ [20-30]}} & \multicolumn{2}{c|}{\textbf{len $\in$ [30-40]}} & \multicolumn{2}{c}{\textbf{len $\in$ [10-20]}} & \multicolumn{2}{c}{\textbf{len $\in$ [20-30]}} & \multicolumn{2}{c}{\textbf{len $\in$ [30-40]}} \\
\cmidrule(lr){2-3} \cmidrule(lr){4-5} \cmidrule(lr){6-7} \cmidrule(lr){8-9} \cmidrule(lr){10-11} \cmidrule(lr){12-13}
 & \textbf{no-CoT} & \textbf{CoT} & \textbf{no-CoT} & \textbf{CoT} & \textbf{no-CoT} & \textbf{CoT} & \textbf{no-CoT} & \textbf{CoT} & \textbf{no-CoT} & \textbf{CoT} & \textbf{no-CoT} & \textbf{CoT} \\
\midrule
\textbf{pure string BPE tokens (a)} &\colorcellgreen{30.10}{20}{99}30.10& \colorcellgreen{45.70}{20}{99}45.70& \colorcellred{15.10}{0}{20}15.10 & \colorcellred{9.10}{0}{20}9.10 & \colorcellred{6.40}{0}{20}6.40 & \colorcellred{2.00}{0}{20}2.00 & \colorcellgreen{33.20}{20}{99}33.20 & \colorcellgreen{47.70}{20}{99}47.70 & \colorcellred{14.00}{0}{20}14.00 & \colorcellred{9.40}{0}{20}9.40 & \colorcellred{3.80}{0}{20}3.80 & \colorcellred{2.70}{0}{20}2.70 \\
\textbf{\texttt{" "}-deliminated token (b)} & \colorcellgreen{46.20}{20}{99}46.20 &\colorcellgreen{58.40}{20}{99}58.40 & \colorcellred{16.10}{0}{20}16.10 & \colorcellgreen{24.90}{20}{99}24.90 & \colorcellred{7.50}{0}{20}7.50 & \colorcellred{10.90}{0}{20}10.90 & \colorcellgreen{45.90}{20}{99}45.90 & \colorcellgreen{63.70}{20}{99}63.70 & \colorcellred{17.60}{0}{20}17.60 & \colorcellgreen{34.00}{20}{99}34.00 & \colorcellred{5.60}{0}{20}5.60 & \colorcellred{18.60}{0}{20}18.60 \\
\textbf{\texttt{", "}-deliminated token (c)} & \colorcellgreen{56.00}{20}{99}56.00 & \colorcellgreen{55.40}{20}{99}55.40 & \colorcellred{19.40}{0}{20}19.40 & \colorcellgreen{38.60}{20}{99}38.60 & \colorcellred{10.20}{0}{20}10.20 & \colorcellgreen{28.10}{20}{99}28.10 & \colorcellgreen{63.60}{20}{99}63.60 & \colorcellgreen{69.30}{20}{99}69.30 & \colorcellgreen{32.80}{20}{99}32.80 & \colorcellgreen{56.10}{20}{99}56.10 & \colorcellred{13.90}{0}{20}13.90 & \colorcellgreen{42.30}{20}{99}42.30 \\
\textbf{precise-item token (d)} & \colorcellgreen{50.70}{20}{99}50.70 & \colorcellgreen{96.80}{20}{99}96.80& \colorcellred{15.80}{0}{20}15.80 & \colorcellgreen{81.60}{20}{99}81.60 & \colorcellred{7.90}{0}{20}7.90 & \colorcellgreen{56.10}{20}{99}56.10 & \colorcellgreen{58.30}{20}{99}58.30 & \colorcellgreen{96.50}{20}{99}96.50 & \colorcellgreen{30.20}{20}{99}30.20 & \colorcellgreen{90.00}{20}{99}90.00 & \colorcellred{12.60}{0}{20}12.60 & \colorcellgreen{70.80}{20}{99}70.80 \\
 \textbf{$\Delta_\text{tok}$  [max] } & {\color{blue} 25.90} &{\color{blue}{41.10}}& {\color{blue} 4.30}&{\color{blue}72.50}&{\color{blue}3.80}&{\color{blue}54.10}&{\color{blue}30.40}&{\color{blue}48.80}&{\color{blue}18.80}&{\color{blue}80.60}&{\color{blue}10.10}&{\color{blue}68.10}
\\
\bottomrule
\end{tabular}
}
\caption{Results of counting \texttt{a}s and \texttt{b}s in string consisting of letter \texttt{a} and \texttt{b}, using GPT-4o-mini API. Numbers indicate the average accuracy (\%) over 1000 random generated instances.  }
\label{tab:main_result_gpt}
\end{table*}

\begin{table*}[t]
\centering
\resizebox{\textwidth}{!}{
\begin{tabular}{rcccccc|cccccc}
\toprule
 & \multicolumn{6}{c|}{\textbf{Counting letter \texttt{e}}} & \multicolumn{6}{c}{\textbf{Counting letter \texttt{z}}} \\
\cmidrule(lr){2-7} \cmidrule(lr){8-13}
 \multirow{2}{*}{\textbf{String-Token Type \ \ \ }}& \multicolumn{2}{c}{\textbf{len $\in$ [10-20]}} & \multicolumn{2}{c}{\textbf{len $\in$ [20-30]}} & \multicolumn{2}{c|}{\textbf{len $\in$ [30-40]}} & \multicolumn{2}{c}{\textbf{len $\in$ [10-20]}} & \multicolumn{2}{c}{\textbf{len $\in$ [20-30]}} & \multicolumn{2}{c}{\textbf{len $\in$ [30-40]}} \\
\cmidrule(lr){2-3} \cmidrule(lr){4-5} \cmidrule(lr){6-7} \cmidrule(lr){8-9} \cmidrule(lr){10-11} \cmidrule(lr){12-13}
 & \textbf{no-CoT} & \textbf{CoT} & \textbf{no-CoT} & \textbf{CoT} & \textbf{no-CoT} & \textbf{CoT} & \textbf{no-CoT} & \textbf{CoT} & \textbf{no-CoT} & \textbf{CoT} & \textbf{no-CoT} & \textbf{CoT} \\
\midrule
\textbf{pure string BPE tokens (a)} & \colorcellgreen{26.60}{20}{99}26.60 & \colorcellgreen{55.20}{20}{99}55.20 & \colorcellred{19.80}{0}{20}19.80 & \colorcellred{12.20}{0}{20}12.20 & \colorcellred{11.40}{0}{20}11.40 & \colorcellred{2.10}{0}{20}2.10 & \colorcellgreen{31.10}{20}{99}31.10 & \colorcellgreen{59.10}{20}{99}59.10 & \colorcellred{11.70}{0}{20}11.70 & \colorcellgreen{22.10}{20}{99}22.10 & \colorcellred{4.60}{0}{20}4.60 & \colorcellred{7.30}{0}{20}7.30 \\
\textbf{\texttt{" "}-deliminated token (b)} & \colorcellgreen{41.00}{20}{99}41.00 & \colorcellgreen{52.90}{20}{99}52.90 & \colorcellgreen{23.90}{20}{99}23.90 & \colorcellgreen{28.20}{20}{99}28.20 & \colorcellred{13.00}{0}{20}13.00 & \colorcellred{16.00}{0}{20}16.00 & \colorcellgreen{45.30}{20}{99}45.30 & \colorcellgreen{63.99}{20}{99}63.90 & \colorcellred{16.60}{0}{20}16.60 & \colorcellgreen{46.20}{20}{99}46.20 & \colorcellred{6.80}{0}{20}6.80 & \colorcellgreen{29.50}{20}{99}29.50 \\
\textbf{\texttt{", "}-deliminated token (c)} & \colorcellgreen{45.50}{20}{99}45.50 & \colorcellgreen{64.20}{20}{99}64.20 & \colorcellgreen{27.40}{20}{99}27.40 & \colorcellgreen{44.20}{20}{99}44.20 & \colorcellred{18.00}{0}{20}18.00 & \colorcellgreen{27.60}{20}{99}27.60 & \colorcellgreen{56.20}{20}{99}56.20 & \colorcellgreen{73.60}{20}{99}73.60 & \colorcellgreen{28.20}{20}{99}28.20 & \colorcellgreen{55.60}{20}{99}55.60 & \colorcellred{13.90}{0}{20}13.90 & \colorcellgreen{41.90}{20}{99}41.90 \\
\textbf{precise-item token (d)} & \colorcellgreen{60.10}{20}{99}60.10 & \colorcellgreen{97.70}{20}{99}97.70 & \colorcellgreen{32.50}{20}{99}32.50 & \colorcellgreen{89.30}{0}{99}89.30 & \colorcellred{15.30}{0}{20}15.30 & \colorcellgreen{70.70}{0}{99}70.70 & \colorcellgreen{60.60}{20}{99}60.60 & \colorcellgreen{98.40}{20}{99}98.40 & \colorcellgreen{30.60}{20}{99}30.60 & \colorcellgreen{93.80}{20}{99}93.80 & \colorcellred{13.30}{0}{20}13.30 & \colorcellgreen{74.80}{20}{99}74.80 \\
\textbf{$\Delta_\text{tok}$  [max] } &{\color{blue}33.50}&{\color{blue}44.80}&{\color{blue}12.70}&{\color{blue}77.10}&{\color{blue}6.60}&{\color{blue}68.60}&{\color{blue}29.50}&{\color{blue}39.30}&{\color{blue}18.90}&{\color{blue}71.70}&{\color{blue}9.30}&{\color{blue}67.50}\\
\bottomrule
\end{tabular}
}
\caption{Results of counting \texttt{e}s and \texttt{z}s in string consisting of letter \texttt{e} and \texttt{z}, using GPT-4o-mini model. Numbers indicate the average accuracy (\%) over 1000 random generated instances.}
\label{table:ez}
\end{table*}

\begin{table}[t]
\centering
\small
\renewcommand{\arraystretch}{1.1}
\resizebox{\columnwidth}{!}{
\begin{tabular}{c c ccc ccc ccc}
\toprule[1.5pt]
\multicolumn{2}{c}{\multirow{2}{*}{\textbf{Method/Length}}} & \multicolumn{3}{c}{\textbf{Letter}} & \multicolumn{3}{c}{\textbf{Letter+Digit}} & \multicolumn{3}{c}{\textbf{Digit}} \\
\cmidrule(lr){3-5} \cmidrule(lr){6-8} \cmidrule(lr){9-11}
\multicolumn{2}{c}{} & \textbf{Str} & \textbf{List} & \textbf{$\Delta_\text{tok}$}  & \textbf{Str} & \textbf{List} & \textbf{$\Delta_\text{tok}$} &  \textbf{Str} & \textbf{List} & \textbf{$\Delta_\text{tok}$} \\
\midrule
\multirow{3}{*}{\textbf{CoT}} & \textbf{5-10} & 24.6 & \textbf{32.0}&\color{blue}7.4 & 30.1 & \textbf{35.7}&\color{blue}5.6 & 56.2 & \textbf{84.4}&\color{blue}27.8 \\
 & \textbf{10-15} & 3.3 & 8.7&\color{blue}5.4 & 5.4 & \textbf{10.1}&\color{blue}4.7 & 7.9 & \textbf{33.0}&\color{blue}25.1 \\
 & \textbf{15-20} & 0.4 & 1.1&\color{blue}0.7 & 0.7 & 2.7&\color{blue}2.0 & 0.8 & 4.8&\color{blue}4.0 \\
\midrule
\multirow{3}{*}{\textbf{SCoT}} & \textbf{5-10} & 28.0 & \textbf{35.1}&\color{blue}7.1 & 31.1 & \textbf{38.5}&\color{blue}7.4 & 64.7 & \textbf{84.8}&\color{blue}20.1 \\
 & \textbf{10-15} & 10.4 & \textbf{12.6}&\color{blue}2.2 & 10.6 & \textbf{15.6}&\color{blue}5.0 & 15.6 & \textbf{34.6}&\color{blue}19.0 \\
 & \textbf{15-20} & 2.4 & 3.4&\color{blue}1.0& 2.6 & 4.7 &\color{blue}2.1& 3.2 & 8.6&\color{blue}5.4 \\
\bottomrule[1.5pt]
\end{tabular}
}
\caption{Performance on \textbf{ sorting tasks } using GPT-4o mini with Chain-of-Thought (CoT) and Supervised Chain-of-Thought (SCoT) across different input types, length ranges and tokenization types. }
\label{tab:sort}
\end{table}

\begin{table}[t]
\centering
\small
\renewcommand{\arraystretch}{1.1}
\resizebox{0.9\columnwidth}{!}{
\begin{tabular}{c c ccc ccc ccc}
\toprule[1.5pt]
\multicolumn{2}{c}{\multirow{2}{*}{\textbf{Method/Length}}} & \multicolumn{3}{c}{\textbf{Random}} & \multicolumn{3}{c}{\textbf{Word}} & \multicolumn{3}{c}{\textbf{High-freq Word}} \\
\cmidrule(lr){3-5} \cmidrule(lr){6-8} \cmidrule(lr){9-11}
\multicolumn{2}{c}{} & \textbf{Str} & \textbf{List} & \textbf{$\Delta_\text{tok}$} & \textbf{Str} & \textbf{List} &\textbf{$\Delta_\text{tok}$} & \textbf{Str} & \textbf{List}&\textbf{$\Delta_\text{tok}$}  \\
\midrule
\multirow{5}{*}{\textbf{CoT}} & \textbf{5-10} & 46.0 & \textbf{70.0}& {\color{blue}24.0} & 39.1 & \textbf{56.5}& {\color{blue}17.4} & 54.2 & \textbf{66.6}& {\color{blue}12.4} \\
 & \textbf{10-15} & 8.6 & \textbf{38.1}&{\color{blue}29.5} & 11.7 & \textbf{22.4}&{\color{blue}10.7} & 13.6 & \textbf{25.9}& {\color{blue}12.3} \\
 & \textbf{15-20} & 2.5 & \textbf{20.1}& {\color{blue}17.6} & 1.5 & \textbf{8.0}& {\color{blue}6.5} & 2.3 & \textbf{9.0} & {\color{blue}6.70}\\
 & \textbf{20-25} & 0.3 & \textbf{9.6} &{\color{blue}9.3}& 0.5 & \textbf{2.2} & {\color{blue}1.7} & 0.6 & \textbf{2.4} & {\color{blue}1.8} \\
 & \textbf{25-30} & 0.4 & \textbf{4.7} & {\color{blue}4.3}& 0.1 & \textbf{0.7} & {\color{blue}0.6}& \textbf{0.5} & 0.4 & {\color{blue}0.1} \\
\midrule
\multirow{5}{*}{\textbf{SCoT}} & \textbf{5-10} & 50.2 & \textbf{72.1}& {\color{blue}21.9} & 51.1 & \textbf{68.0}& {\color{blue}16.9} & 59.1 & \textbf{72.3}& {\color{blue}23.2} \\
 & \textbf{10-15} & 35.8 & \textbf{56.9}& {\color{blue}21.1} & 29.9 & \textbf{52.8}&\color{blue}22.9 & 33.6 & \textbf{56.5}&\color{blue}22.9 \\
 & \textbf{15-20} & 24.9 & \textbf{44.4}&\color{blue}19.5 & 18.7 & \textbf{38.9}&\color{blue}20.2 & 21.5 & \textbf{43.9}&\color{blue}22.4 \\
 & \textbf{20-25} & 18.6 & \textbf{31.3}&\color{blue}12.7 & 13.6 & \textbf{30.1}&\color{blue}16.5 & 12.3 & \textbf{32.6}&\color{blue}20.3 \\
 & \textbf{25-30} & 12.6 & \textbf{23.8} &\color{blue}11.2& 7.2 & \textbf{21.4}&\color{blue}14.2 & 8.5 & \textbf{23.8} & \color{blue}15.3\\
\bottomrule[1.5pt]
\end{tabular}
}
\caption{Performance comparison on \textbf{reversing tasks}, using simlar settings as Table~\ref{tab:sort}.}
\label{tab:reverse}
\end{table}

\section{Experiments} 
\subsection{Settings}
We evaluate the impact of tokenization and input formatting on symbolic reasoning capabilities of large language models (LLMs). We focus on three fundamental symbolic tasks: Arithmetic counting, Sorting and Sorting. 

Each task operates on inputs drawn from controlled domains (letters, digits, or words), and varies in input length and tokenization strategy. Despite differing surface forms, all tasks share a symbolic core that requires composition, memory, and manipulation of atomic units. Importantly, the \textit{task identity} remains unchanged across formatting conditions.

To isolate the role of tokenization, we disable tool use in all models and treat LLMs as black-box functions $\mathcal{M} \circ \mathcal{T}$, where $\mathcal{T}$ is the tokenizer and $\mathcal{M}$ is the model.

For counting tasks, we test four competitive LLMs: GPT-4o-mini, Claude 3.5 Sonnet, Qwen Turbo, and OpenAI o1. For sorting and reversing, we focus on GPT-4o-mini due to its consistent performance and API accessibility.
Each experiment uses 1,000 randomly generated input instances per length bucket. Input lengths are task-specific: for counting, we use lengths in $[10, 20]$, $[20, 30]$, and $[30, 40]$; for sorting, lengths are in $[5, 10]$, $[10, 15]$; and for reversing, in $[5, 30]$ (5-step increments). All models use identical prompts. We evaluate with and without CoT reasoning, and for some tasks apply supervised CoT (SCoT)~\cite{zhang2025doescotpromptnot} to control for CoT quality. Evaluation measures exact-match accuracy.

\subsection{Tokenizer Sensitivity in Symbolic Tasks}

Across all tasks, we observe a consistent phenomenon: model performance varies dramatically with tokenization format, even when the underlying symbolic function $f$ remains fixed. Following are detailed analysis across all tasks conducted:
\begin{figure*}[t]

    \centering
    \includegraphics[width=0.7\linewidth]{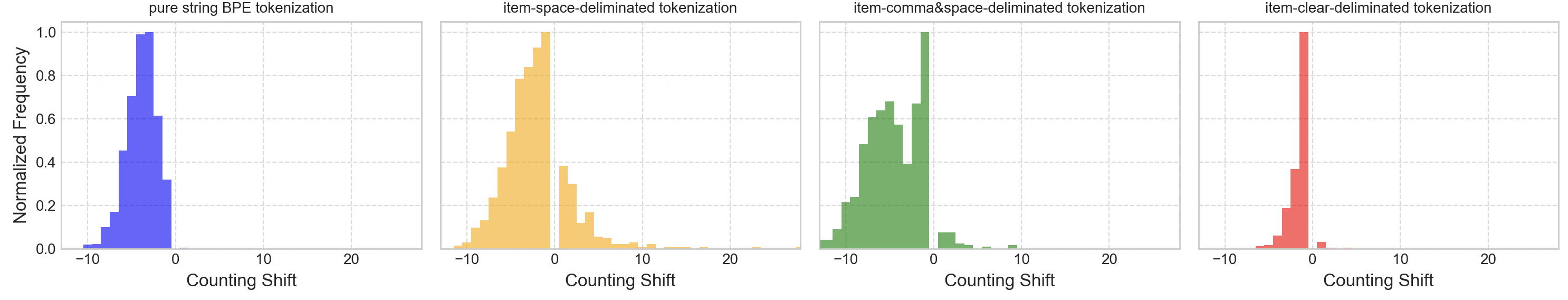}
    
    \caption{Distribution of shifts from the correct count.}
    \label{fig:shift}
\vspace{-1em}
\end{figure*}

\subsubsection*{CoT grants compositional logic}

Enabling Chain-of-Thought (CoT) significantly boosts performance, particularly for tasks that require sequential or compositional logic. This improvement is most pronounced when input length increases, suggesting that CoT enables models to simulate recurrent computation by externalizing intermediate state. In tasks where CoT is not used, performance plateaus or drops sharply as sequence length grows, reflecting the fixed-depth constraint of answer-only Transformers.

\subsubsection*{Symbolic Reasoning is sensitive to input token structure }
Experiments show that tokenization plays a critical role in determining symbolic generalization. For a fixed task function $f$, changes in tokenization alone—without altering task semantics—can yield over 70\% variance in accuracy. This phenomenon is captured quantitatively by the tokenization degradation gap $\Delta_{\text{tok}}$, which consistently reaches high values across all experiments. In Table \ref{tab:main_result_gpt}, for example, switching from raw BPE inputs (type a) to atomic-aligned inputs (type d) improves accuracy by $\Delta_{\text{tok}} = 54.1\%$ for counting \texttt{a}, and similar gains are observed in Table \ref{table:ez} for letters \texttt{e} and \texttt{z}.

The results in Tables \ref{tab:reverse} and \ref{tab:sort} further reveal that symbolic reasoning ability is not only sensitive to tokenizer (controlled by using string vs. list), but also to the type of atomic unit being processed. Tasks involving digits consistently yield significantly higher performance than those involving letters or words, even when the overall structure of the task and input formatting are matched.

For instance, in sorting tasks (Table \ref{tab:sort}), CoT performance on digit sequences reaches up to 84.8\% accuracy for lengths 5–10 in list format, with a corresponding $\Delta_{\text{tok}}$ of 27.8. In contrast, performance drops to 35.1\% on letter sequences under identical conditions, with a much smaller $\Delta_{\text{tok}}$ of 7.1. This discrepancy persists across input lengths and holds under both CoT and SCoT prompting. A similar trend is observed in reversing tasks (Table \ref{tab:reverse}): digit and high-frequency word sequences achieve the highest absolute accuracies and largest gains from structured formatting, suggesting that token content—i.e., whether the model processes compact numerical symbols or open-vocabulary lexical tokens—has a significant effect on symbolic generalization.

These results validate our theoretical claim that symbolic reasoning over atomic units cannot emerge reliably unless tokenization preserves unit-level structure. When input tokens merge multiple semantic units (e.g., letters or digits), the model cannot apply symbolic operations like comparison or increment at the proper resolution. This leads to brittle reasoning and reliance on memorization. In contrast, atomic-aligned formats ensure that symbolic computation is recoverable from token-level patterns, enabling models to generalize even across longer inputs or different domains (letters vs. digits vs. words).

Overall, the combination of CoT and precise tokenization unlocks the model's latent arithmetic and symbolic capabilities. However, without either component, performance degrades—even if the model architecture is unchanged. These findings confirm that CoT grants access to general symbolic reasoning under ideal token granularity, and that a high $\Delta_{\text{tok}}$ is a strong indicator that a model’s performance hinges on token alignment rather than true generalization.

\subsection{Error Shifts Reveal BPE-Induced Counting Failures}

We define error shifts as the difference between the model-predicted and true counts on failed instances. As shown in Figure~\ref{fig:shift}, GPT-4o mini exhibits a strong bias toward negative shifts across all tokenization types, indicating systematic undercounting. With pure BPE tokenization, shifts are exclusively negative—likely due to the model’s inability to parse individual characters within merged tokens (e.g., "abaa"), often resulting in zero counts for target symbols (see Appendix Figure~\ref{fig:tokenlongall}).

When delimiter-separated formats (types (b)–(d)) are used, some positive shifts appear, likely caused by overcounting or retrieval inconsistencies. Yet with fully atomic-aligned tokens (type (d)), errors narrow to a band between –1 and –3, reflecting smaller arithmetic missteps rather than structural confusion. This confirms that BPE introduces larger, systematic errors, whereas cleaner tokenization mitigates extreme deviations.

\section{Conclusion}
We have demonstrated that tokenization is a critical bottleneck in the symbolic reasoning ability of language models. Even with Chain-of-Thought prompting, coarse or misaligned token structures prevent models from accurately performing arithmetic and structured symbolic tasks. Our theoretical framework and empirical findings jointly show that both token format and token type (e.g., digits vs. letters) significantly affect generalization. Aligning tokenization with atomic reasoning units enables smaller models to rival or surpass larger ones, highlighting the need to treat tokenization design as a core component of model capabilities—not merely a preprocessing step.

\section*{Limitations}
Our experiments were conducted on GPT-4o Mini, Claude 3.5 Sonnet and Qwen Turbo. While all models demonstrated strong patterns and consistent evidence showing that certain types of tokenization significantly improve counting performance, we did not extend our testing to other open-source LLMs such as LLaMA, Mistral. This was primarily due to budget and time constraints, as well as preliminary findings that these models exhibited weaker instruction-following abilities compared to GPT and Claude, making the evaluation process more challenging. However, we believe our research remains robust despite these limitations, as mainstream model training and design principles are largely universal, and the patterns observed are likely generalizable to other LLMs.

Additionally, our experiments did not explore extreme context lengths, such as counting instances with more than several hundred tokens. We found that such cases often led to instability due to the accumulation of long CoT steps. We aim to further investigate this aspect as LLMs improve in handling long-context retrieval and generation.
\bibliography{custom}

\begin{thebibliography}{32}
\expandafter\ifx\csname natexlab\endcsname\relax\def\natexlab#1{#1}\fi

\bibitem[{Achiam et~al.(2023)Achiam, Adler, Agarwal, Ahmad, Akkaya, Aleman, Almeida, Altenschmidt, Altman, Anadkat et~al.}]{achiam2023gpt}
Josh Achiam, Steven Adler, Sandhini Agarwal, Lama Ahmad, Ilge Akkaya, Florencia~Leoni Aleman, Diogo Almeida, Janko Altenschmidt, Sam Altman, Shyamal Anadkat, et~al. 2023.
\newblock Gpt-4 technical report.
\newblock \emph{arXiv preprint arXiv:2303.08774}.

\bibitem[{Ackerman and Cybenko(2020)}]{ackerman2020survey}
Joshua Ackerman and George Cybenko. 2020.
\newblock A survey of neural networks and formal languages.
\newblock \emph{arXiv preprint arXiv:2006.01338}.

\bibitem[{Allen-Zhu and Li(2023)}]{allen2023physics}
Zeyuan Allen-Zhu and Yuanzhi Li. 2023.
\newblock Physics of language models: Part 3.1, knowledge storage and extraction.
\newblock \emph{arXiv preprint arXiv:2309.14316}.

\bibitem[{Bai et~al.(2023)Bai, Bai, Chu, Cui, Dang, Deng, Fan, Ge, Han, Huang et~al.}]{bai2023qwen}
Jinze Bai, Shuai Bai, Yunfei Chu, Zeyu Cui, Kai Dang, Xiaodong Deng, Yang Fan, Wenbin Ge, Yu~Han, Fei Huang, et~al. 2023.
\newblock Qwen technical report.
\newblock \emph{arXiv preprint arXiv:2309.16609}.

\bibitem[{Boolos et~al.(2002)Boolos, Burgess, and Jeffrey}]{boolos2002computability}
George~S Boolos, John~P Burgess, and Richard~C Jeffrey. 2002.
\newblock \emph{Computability and logic}.
\newblock Cambridge university press.

\bibitem[{Chang and Bisk(2024)}]{chang2024language}
Yingshan Chang and Yonatan Bisk. 2024.
\newblock Language models need inductive biases to count inductively.
\newblock \emph{arXiv preprint arXiv:2405.20131}.

\bibitem[{Cooper(2017)}]{cooper2017computability}
S~Barry Cooper. 2017.
\newblock \emph{Computability theory}.
\newblock Chapman and Hall/CRC.

\bibitem[{De~Bruijn(1964)}]{de1964polya}
Nicolaas~Govert De~Bruijn. 1964.
\newblock Polya's theory of counting.
\newblock In \emph{Applied combinatorical mathematics}, pages 144--184. Eindhoven University of Technology.

\bibitem[{Del{\'e}tang et~al.(2022)Del{\'e}tang, Ruoss, Grau-Moya, Genewein, Wenliang, Catt, Cundy, Hutter, Legg, Veness et~al.}]{deletang2022neural}
Gr{\'e}goire Del{\'e}tang, Anian Ruoss, Jordi Grau-Moya, Tim Genewein, Li~Kevin Wenliang, Elliot Catt, Chris Cundy, Marcus Hutter, Shane Legg, Joel Veness, et~al. 2022.
\newblock Neural networks and the chomsky hierarchy.
\newblock \emph{arXiv preprint arXiv:2207.02098}.

\bibitem[{Devlin(2018)}]{devlin2018bert}
Jacob Devlin. 2018.
\newblock Bert: Pre-training of deep bidirectional transformers for language understanding.
\newblock \emph{arXiv preprint arXiv:1810.04805}.

\bibitem[{Feng et~al.(2024)Feng, Zhang, Gu, Ye, He, and Wang}]{feng2024towards}
Guhao Feng, Bohang Zhang, Yuntian Gu, Haotian Ye, Di~He, and Liwei Wang. 2024.
\newblock Towards revealing the mystery behind chain of thought: a theoretical perspective.
\newblock \emph{Advances in Neural Information Processing Systems}, 36.

\bibitem[{Fischer et~al.(1968)Fischer, Meyer, and Rosenberg}]{fischer1968counter}
Patrick~C Fischer, Albert~R Meyer, and Arnold~L Rosenberg. 1968.
\newblock Counter machines and counter languages.
\newblock \emph{Mathematical systems theory}, 2(3):265--283.

\bibitem[{Gregor et~al.(2014)Gregor, Danihelka, Mnih, Blundell, and Wierstra}]{gregor2014deep}
Karol Gregor, Ivo Danihelka, Andriy Mnih, Charles Blundell, and Daan Wierstra. 2014.
\newblock Deep autoregressive networks.
\newblock In \emph{International Conference on Machine Learning}, pages 1242--1250. PMLR.

\bibitem[{Gu and Dao(2023)}]{gu2023mamba}
Albert Gu and Tri Dao. 2023.
\newblock Mamba: Linear-time sequence modeling with selective state spaces.
\newblock \emph{arXiv preprint arXiv:2312.00752}.

\bibitem[{Ibarra et~al.(2002)Ibarra, Su, Dang, Bultan, and Kemmerer}]{ibarra2002counter}
Oscar~H Ibarra, Jianwen Su, Zhe Dang, Tevfik Bultan, and Richard~A Kemmerer. 2002.
\newblock Counter machines and verification problems.
\newblock \emph{Theoretical Computer Science}, 289(1):165--189.

\bibitem[{Jerrum(1995)}]{jerrum1995computational}
Mark Jerrum. 1995.
\newblock The computational complexity of counting.
\newblock In \emph{Proceedings of the International Congress of Mathematicians: August 3--11, 1994 Z{\"u}rich, Switzerland}, pages 1407--1416. Springer.

\bibitem[{Li et~al.(2024{\natexlab{a}})Li, Liu, Zhou, and Ma}]{li2024chain}
Zhiyuan Li, Hong Liu, Denny Zhou, and Tengyu Ma. 2024{\natexlab{a}}.
\newblock Chain of thought empowers transformers to solve inherently serial problems.
\newblock \emph{arXiv preprint arXiv:2402.12875}.

\bibitem[{Li et~al.(2024{\natexlab{b}})Li, Liu, Zhou, and Ma}]{li2024chainthoughtempowerstransformers}
Zhiyuan Li, Hong Liu, Denny Zhou, and Tengyu Ma. 2024{\natexlab{b}}.
\newblock \href {http://arxiv.org/abs/2402.12875} {Chain of thought empowers transformers to solve inherently serial problems}.

\bibitem[{Liu et~al.(2022)Liu, Zhang, and Mou}]{liu2022character}
Puyuan Liu, Xiang Zhang, and Lili Mou. 2022.
\newblock A character-level length-control algorithm for non-autoregressive sentence summarization.
\newblock \emph{Advances in Neural Information Processing Systems}, 35:29101--29112.

\bibitem[{Peng et~al.(2023)Peng, Alcaide, Anthony, Albalak, Arcadinho, Biderman, Cao, Cheng, Chung, Grella et~al.}]{peng2023rwkv}
Bo~Peng, Eric Alcaide, Quentin Anthony, Alon Albalak, Samuel Arcadinho, Stella Biderman, Huanqi Cao, Xin Cheng, Michael Chung, Matteo Grella, et~al. 2023.
\newblock Rwkv: Reinventing rnns for the transformer era.
\newblock \emph{arXiv preprint arXiv:2305.13048}.

\bibitem[{Rodriguez et~al.(1999)Rodriguez, Wiles, and Elman}]{rodriguez1999recurrent}
Paul Rodriguez, Janet Wiles, and Jeffrey~L Elman. 1999.
\newblock A recurrent neural network that learns to count.
\newblock \emph{Connection Science}, 11(1):5--40.

\bibitem[{Rosenblatt(1958)}]{rosenblatt1958perceptron}
Frank Rosenblatt. 1958.
\newblock The perceptron: a probabilistic model for information storage and organization in the brain.
\newblock \emph{Psychological review}, 65(6):386.

\bibitem[{Sanford et~al.(2024)Sanford, Hsu, and Telgarsky}]{sanford2024transformers}
Clayton Sanford, Daniel Hsu, and Matus Telgarsky. 2024.
\newblock Transformers, parallel computation, and logarithmic depth.
\newblock \emph{arXiv preprint arXiv:2402.09268}.

\bibitem[{Sennrich(2015)}]{sennrich2015neural}
Rico Sennrich. 2015.
\newblock Neural machine translation of rare words with subword units.
\newblock \emph{arXiv preprint arXiv:1508.07909}.

\bibitem[{Suzgun et~al.(2019)Suzgun, Gehrmann, Belinkov, and Shieber}]{suzgun2019lstmnetworksperformdynamic}
Mirac Suzgun, Sebastian Gehrmann, Yonatan Belinkov, and Stuart~M. Shieber. 2019.
\newblock \href {http://arxiv.org/abs/1906.03648} {Lstm networks can perform dynamic counting}.

\bibitem[{Touvron et~al.(2023)Touvron, Lavril, Izacard, Martinet, Lachaux, Lacroix, Rozi{\`e}re, Goyal, Hambro, Azhar et~al.}]{touvron2023llama}
Hugo Touvron, Thibaut Lavril, Gautier Izacard, Xavier Martinet, Marie-Anne Lachaux, Timoth{\'e}e Lacroix, Baptiste Rozi{\`e}re, Naman Goyal, Eric Hambro, Faisal Azhar, et~al. 2023.
\newblock Llama: Open and efficient foundation language models.
\newblock \emph{arXiv preprint arXiv:2302.13971}.

\bibitem[{Vaswani(2017)}]{vaswani2017attention}
A~Vaswani. 2017.
\newblock Attention is all you need.
\newblock \emph{Advances in Neural Information Processing Systems}.

\bibitem[{Wei et~al.(2022)Wei, Wang, Schuurmans, Bosma, Xia, Chi, Le, Zhou et~al.}]{wei2022chain}
Jason Wei, Xuezhi Wang, Dale Schuurmans, Maarten Bosma, Fei Xia, Ed~Chi, Quoc~V Le, Denny Zhou, et~al. 2022.
\newblock Chain-of-thought prompting elicits reasoning in large language models.
\newblock \emph{Advances in neural information processing systems}, 35:24824--24837.

\bibitem[{Weiss et~al.(2018)Weiss, Goldberg, and Yahav}]{weiss2018practicalcomputationalpowerfinite}
Gail Weiss, Yoav Goldberg, and Eran Yahav. 2018.
\newblock \href {http://arxiv.org/abs/1805.04908} {On the practical computational power of finite precision rnns for language recognition}.

\bibitem[{Wynn(1990)}]{wynn1990children}
Karen Wynn. 1990.
\newblock Children's understanding of counting.
\newblock \emph{Cognition}, 36(2):155--193.

\bibitem[{Zhang et~al.(2024)Zhang, Abdul-Mageed, and Lakshmanan}]{zhang2024autoregressivechainthought}
Xiang Zhang, Muhammad Abdul-Mageed, and Laks V.~S. Lakshmanan. 2024.
\newblock \href {http://arxiv.org/abs/2409.09239} {Autoregressive + chain of thought = recurrent: Recurrence's role in language models' computability and a revisit of recurrent transformer}.

\bibitem[{Zhang et~al.(2025)Zhang, Cao, Wei, You, and Ding}]{zhang2025doescotpromptnot}
Xiang Zhang, Juntai Cao, Jiaqi Wei, Chenyu You, and Dujian Ding. 2025.
\newblock \href {http://arxiv.org/abs/2503.10084} {Why does your cot prompt (not) work? theoretical analysis of prompt space complexity, its interaction with answer space during cot reasoning with llms: A recurrent perspective}.

\end{thebibliography}
\bibliographystyle{acl_natbib}

\appendix

\section*{Appendix}

\section{Supervised Chain of Thought}
\label{app:scot}
Naive Chain of Thought (CoT), which uses a generic "think step by step" prompt for all tasks, poses significant challenges for models in determining the correct steps, especially for complex, multi-step reasoning tasks. To mitigate this confounding factor, we follow previous work and employ Supervised CoT~\cite{zhang2025doescotpromptnot}, as the derivation of steps is not the focus of our research and should not affect performance due to incorrect CoT steps. Below, we define Supervised CoT and explain its application in counting tasks.
\subsection{Definition}
The search space for solving a task can be viewed as a combination of the prompt space and the answer space. When instructed to perform tasks step by step, language models must devise a step template which is used to determine the actions at each step. This template is crucial for solving tasks, as it specifies what information is processed and how it is computed at each CoT step. However, for a given task, there are numerous ways to perform a "step-by-step" approach, each computing different elements per step. Finding the optimal set of steps is challenging yet essential, as it directly influences the ability to find solutions in the answer space~\cite{zhang2025doescotpromptnot}.

Supervised CoT provides human supervision in determining the step template. Rather than asking the model to develop its own plan for each step, humans identify the "recurrent" procedure in the computation and explicitly instruct the model to follow a specific step template. This approach allows the CoT to bypass the need to search for optimal steps, focusing instead on finding solutions within the answer space under optimal step guidance.

\begin{figure}[htbp]
    \centering
    \includegraphics[width=0.8\linewidth]{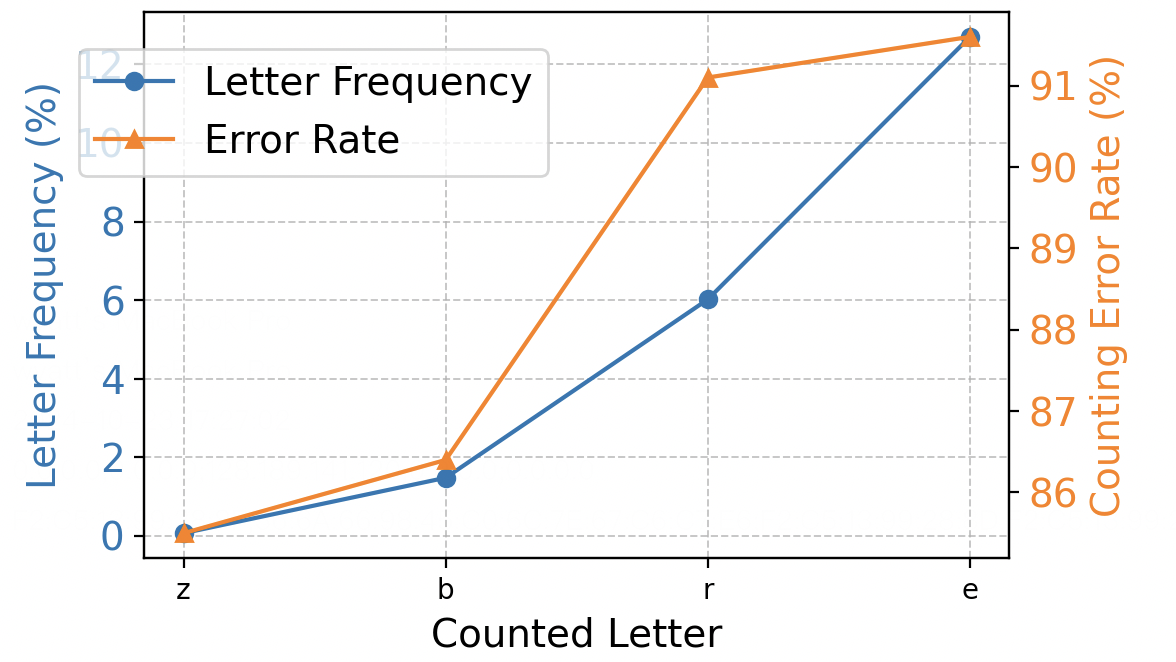}
    \vspace{-0.1em}
    \caption{Counting accuracy (Orange) with respect to target letter frequency (Blue) in Human Natural Language.}
    \label{fig:freq}
    \vspace{0.1em}
\end{figure}

\subsection{Supervised CoT and Counting}
In inductive counting, which relies on CoT to compute the counter value recurrently(Figure \ref{fig:inductive}), it is crucial that each step of CoT accurately extracts and outputs the counter value in text. This output is necessary for the value to be recurrently processed through "string-vector" conversion. Therefore, rather than simply prompting the model with "determine the number of \texttt{a} in the given string" using the generic instruction "think step by step," we specifically instruct the model to print out a counter value at each step. We explicitly define the step template to ensure the model follows the optimal CoT steps, preventing deviations or the use of suboptimal steps.

\begin{table*}[ht]
\centering

\resizebox{0.7\textwidth}{!}{
\begin{tabular}{ccc|cc}
\toprule
\multirow{2}{*}{\textbf{String-token Type}} & \multicolumn{2}{c|}{\textbf{Counting \texttt{a}}} & \multicolumn{2}{c}{\textbf{Counting \texttt{b}}} \\
\cmidrule(lr){2-3} \cmidrule(lr){4-5}
 & \textbf{Unsupervised-CoT} & \textbf{Supervised CoT} & \textbf{Unsupervised-CoT} & \textbf{Supervised CoT} \\
\midrule
\textbf{(b)} & 8.40 & \textbf{10.90} & \textbf{20.70} & 18.60 \\
\textbf{(c)} & 24.00 & \textbf{28.10} & 29.30 & \textbf{42.30} \\
\textbf{(d)} & 34.90 & \textbf{56.10} & 42.70 & \textbf{70.80} \\

\bottomrule
\end{tabular}
}
\caption{Counting experiments in the length range of 30-40 comparing Supervised CoT and Unsupervised CoT. The bolded font indicates the better performance in the pairwise comparison between Supervised and Unsupervised CoT.}
\label{tab:Supervision}
\end{table*}
\noindent\textbf{Experiments.} We demonstrate the significant performance gap between Supervised and Unsupervised CoT. Specifically, we observe that supervision not only helps the model accurately extract the counter but also ensures it follows the correct steps (e.g., an incorrect step would be outputting whether the current letter is the target, rather than extracting the counter value). Even when Unsupervised CoT identifies the correct steps (i.e., extracting the counter into text), we still notice more frequent errors during the extraction process compared to Supervised CoT, which imposes strict constraints on what to extract at each step. The comparison between Supervised and Unsupervised CoT is presented in Table \ref{tab:Supervision}, showing a clear dominance of Supervised CoT, with accuracy gains observed in nearly all cases.

\section{Comprehensive Experiments on the Relationship Between Letter Frequency and Symbolic Reasoning Performance}
\label{app:freq_analysis}
\begin{figure}[t]
    \centering
    \includegraphics[width=0.85\linewidth]{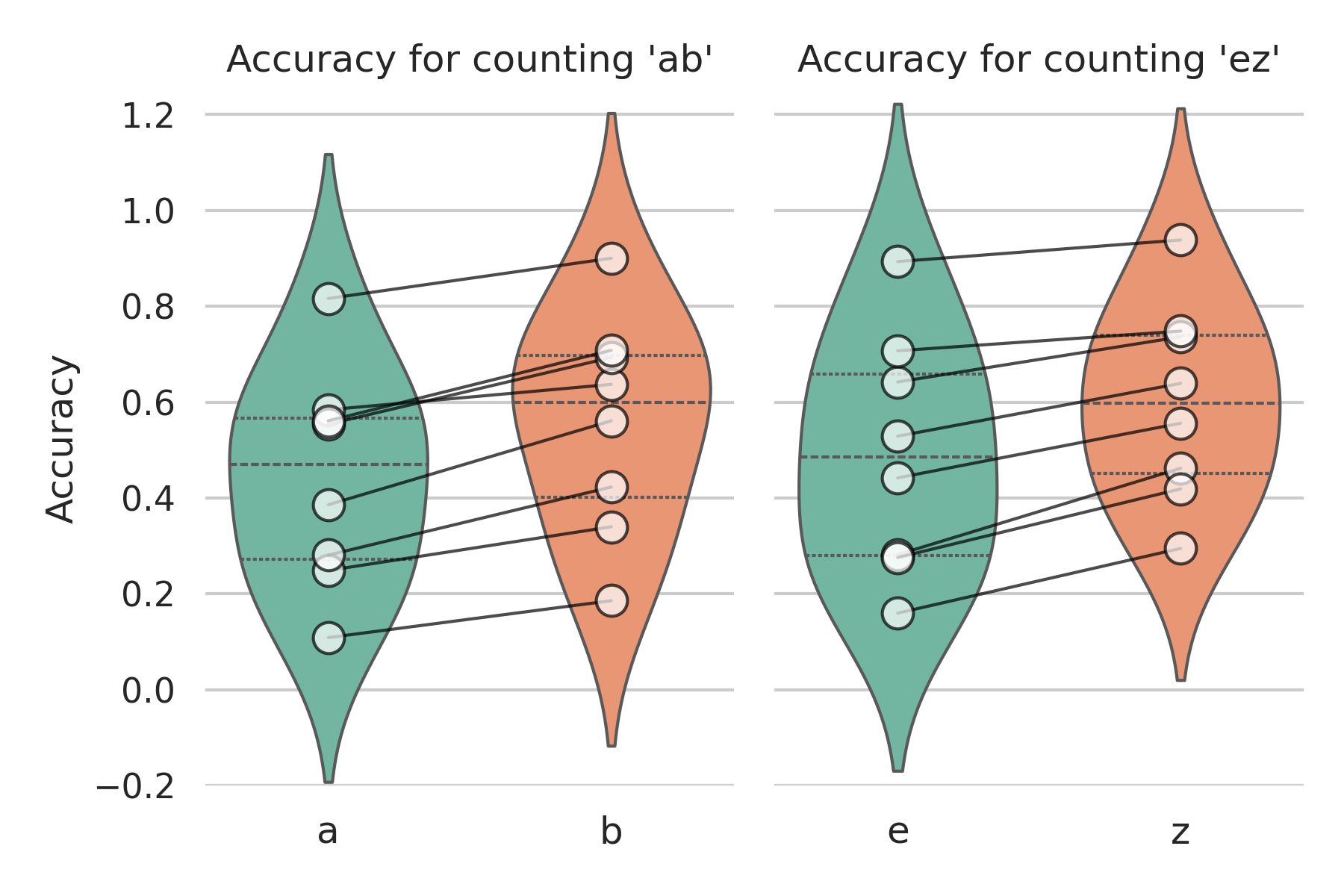}
    \vspace{-0.1em}
    \caption{Pairwise comparison of counting accuracy for different letters in strings. The left plot shows the distribution of accuracy for \texttt{a} and \texttt{b} in \texttt{ab} strings, with each dot representing the average accuracy for \texttt{a} in a given CoT case (e.g., spaced-string in the [10,20] range), connected to the corresponding accuracy for \texttt{b} in the same setting. The right plot illustrates a similar case for \texttt{e} and \texttt{z} in \texttt{ez} strings. Note: The y-axis limit exceeds [0,1] as the distribution is calculated based on variance and mean, with larger variance pushing the upper bound of the confidence interval beyond the maximum value.}
    \label{fig:violin}
    \vspace{-1em}
\end{figure}

Our results in counting experiments show consistently higher counting accuracy for the letter \texttt{b} compared to \texttt{a} across all proper counting settings (CoT enabled, non-BPE tokenization), as shown in Table~\ref{tab:main_result_gpt} and Figure~\ref{fig:violin} left. We hypothesized this difference stems from varying letter frequencies in natural language affecting token-embedding sensitivity.

To further investigate this hypothesis, we compared counting performance between the most frequent letter \texttt{e} (12.7\%) and least frequent letter \texttt{z} (0.07\%) in English. Results in Table \ref{table:ez} show \texttt{z} significantly outperforming \texttt{e}, mirroring the pattern seen with \texttt{b} (1.5\%) versus \texttt{a} (8.2\%). The accuracy advantage for lower-frequency letters ranges from 3-14\% (Figure~\ref{fig:violin}).

Our results reveal that lower-frequency tokens carry less embedded information from training, making them easier to track through the attention mechanism. In contrast, common letters like \texttt{a} and \texttt{e} may encode more complex linguistic information, potentially interfering with counting tasks.

To verify these results beyond the letter pairs \texttt{a}, \texttt{b} and \texttt{e}, \texttt{z}, we selected another set of letters with significantly different frequencies in \textit{human languages}, according to Wikipedia: \texttt{z} (0.07\%), \texttt{b} (1.48\%), \texttt{r} (6.02\%), and \texttt{e} (12.70\%). We generated counting instances of lengths between 80 and 100—ensuring that each letter appears more than 20 times on average—by uniformly sampling one of the four letters to form each string (e.g., \texttt{zrrbeez}). We then performed counting for each letter in the generated strings. As shown in Table \ref{tab:zbre}, a consistent trend was observed across tokenization types (b)–(d) (excluding (a), as pure BPE was previously shown not to yield meaningful counting results). Specifically, rare tokens consistently outperformed more frequent tokens in natural language, with performance improvements ranging from 6\% to 12\%. Figure \ref{fig:freq} visually compares performance and letter frequency, showing an overlap between frequency and error rate. We suspect that rare letters carry less information in their embeddings, reducing distraction during the attention calculation in the counting process.

\begin{table}[t]
\centering
\resizebox{0.5\textwidth}{!}{
\begin{tabular}{c|cccccc}
\toprule
\multirow{1}{*}{\textbf{string-token }} & \multicolumn{2}{c}{\textbf{len $\in$ [10-20]}} & \multicolumn{2}{c}{\textbf{len $\in$ [20-30]}} & \multicolumn{2}{c}{\textbf{len $\in$ [30-40]}} \\
\cmidrule(lr){2-3} \cmidrule(lr){4-5} \cmidrule(lr){6-7}
 \textbf{type} & \textbf{count \texttt{a}} & \textbf{count \texttt{b}} & \textbf{count \texttt{a}} & \textbf{count \texttt{b}} & \textbf{count \texttt{a}} & \textbf{count \texttt{b}} \\
\midrule
\textbf{(a)} & \colorcellgreen{86.30}{80}{100}86.30 & \colorcellgreen{86.32}{80}{100}86.20 & 62.40 & 65.20 & 50.60 & 54.40 \\
\textbf{(b)} & \colorcellgreen{90.60}{80}{100}90.60 & \colorcellgreen{94.00}{80}{100}94.00 & \colorcellgreen{80.40}{80}{100}80.40& \colorcellgreen{87.50}{80}{100}87.50& 76.10 & 79.60\\
\textbf{(c)} & \colorcellgreen{94.90}{80}{100}94.90 & \colorcellgreen{97.70}{80}{100}97.70  & \colorcellgreen{92.80}{80}{100}92.80 & \colorcellgreen{97.90}{80}{100}97.90 & \colorcellgreen{91.40}{80}{100}91.40 & \colorcellgreen{94.20}{80}{100}94.20\\
\textbf{(d)} & \colorcellgreen{93.00}{80}{100}93.00 & \colorcellgreen{94.20}{80}{100}94.20 &\colorcellgreen{87.80}{80}{100}87.80& \colorcellgreen{91.00}{80}{100}91.00& \colorcellgreen{87.30}{80}{100}87.30& \colorcellgreen{89.80}{80}{100}89.80 \\
\bottomrule
\end{tabular}

}
\caption{Counting results on strings with letter \texttt{a} and \texttt{b}, using Claude 3.5 Sonnet API. All results are using supervised CoT~\cite{zhang2025doescotpromptnot}, with same prompt for GPT-4o mini. Numbers indicate the average accuracy (\%) over 1000 random generated instances.}
\label{tab:claude}
\end{table}

\begin{table}[t!]
\centering

\resizebox{0.4\textwidth}{!}{
\begin{tabular}{ccccc}
\toprule
\multirow{2}{*}{\textbf{String-Token Type}} & \multicolumn{4}{c}{\textbf{len $\in$ [80, 100]}} \\
\cmidrule(lr){2-5}
 & \textbf{z} & \textbf{b} & \textbf{r} & \textbf{e} \\
\midrule
\textbf{(b)} & \colorcellgreen{14.50}{8}{15}14.50 & \colorcellgreen{13.60}{8}{15}13.60 & \colorcellgreen{8.90}{8}{15}8.90 & \colorcellgreen{8.40}{8}{15}8.40 \\ \hline
\textbf{(c)} & \colorcellgreen{36}{24}{37}36.00 & \colorcellgreen{36}{24}{37}36.60 & \colorcellgreen{28}{24}{37}28.30 & \colorcellgreen{24}{24}{37}24.30 \\ \hline
\textbf{(d)} & \colorcellgreen{62}{50}{62}61.60 & \colorcellgreen{60}{50}{62}60.20 & \colorcellgreen{53}{50}{62}54.10 & \colorcellgreen{52}{50}{62}51.90 \\ \hline
\multicolumn{5}{c}{\textbf{Letter Frequency}}
\\
percentage & \colorcellgreen{1}{12}{0}0.07 & \colorcellgreen{2}{12}{0}1.48 & \colorcellgreen{6}{12}{0}6.02 & \colorcellgreen{12}{12}{0}12.70 \\ 
\bottomrule
\end{tabular}
}
\caption{Counting performance of letters that have very different letter frequency in human language. }

\label{tab:zbre}
\end{table}

\begin{figure}[htbp]
    \centering
    \includegraphics[width=\linewidth]{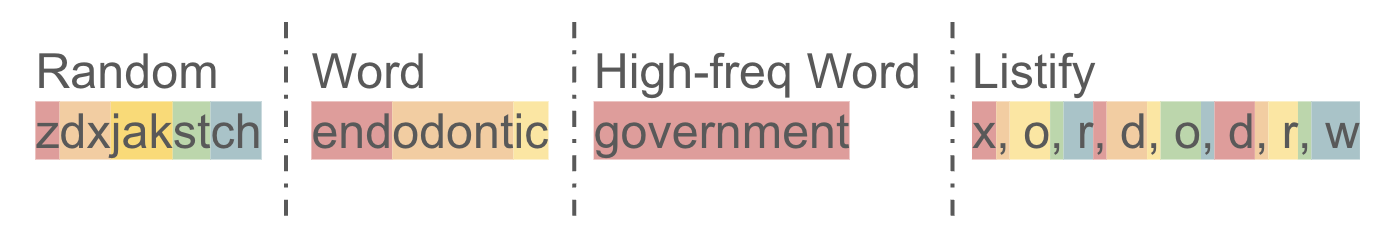}
    \caption{Tokenization patterns of the GPT-4o tokenizer when processing four different input types for reversing task: (a) random character strings, (b) regular dictionary words, (c) high-frequency English words, and (d) listified random strings with explicit delimiters.}
    \label{fig:reverse_token}
    \vspace{0.1em}
\end{figure}

\begin{figure}[htbp]
    \centering
    \includegraphics[width=\linewidth]{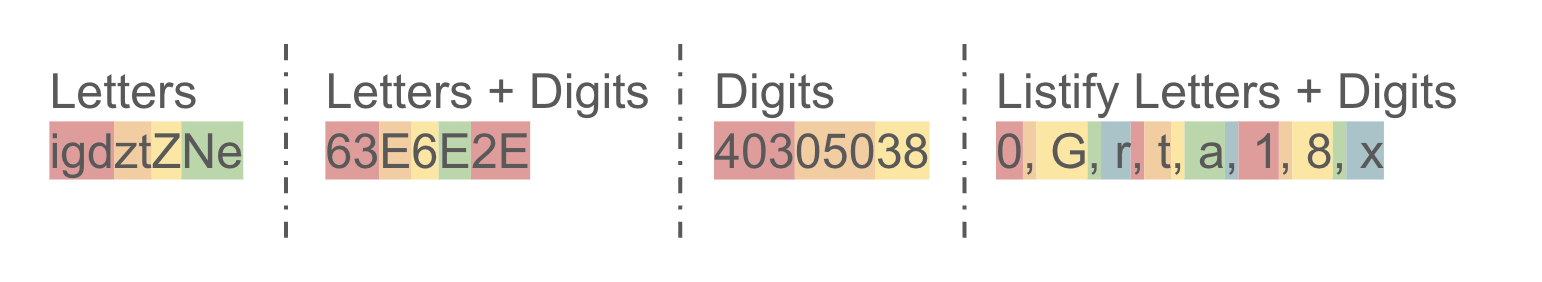}
    \caption{Tokenization patterns of the GPT-4o tokenizer across diverse input compositions for sorting task: (a) random letter strings composed solely of alphabetic characters, (b) mixed random strings containing both letters and digits, (c) random digit strings composed exclusively of numerical characters, and (d) listified mixed strings with explicit delimiters separating letter and digit combinations.}
    \label{fig:sort_token}
    \vspace{0.1em}
\end{figure}

\section{Tokenization in Different LLMs}
\begin{figure*}[htbp]
    \centering
    \includegraphics[width=\linewidth]{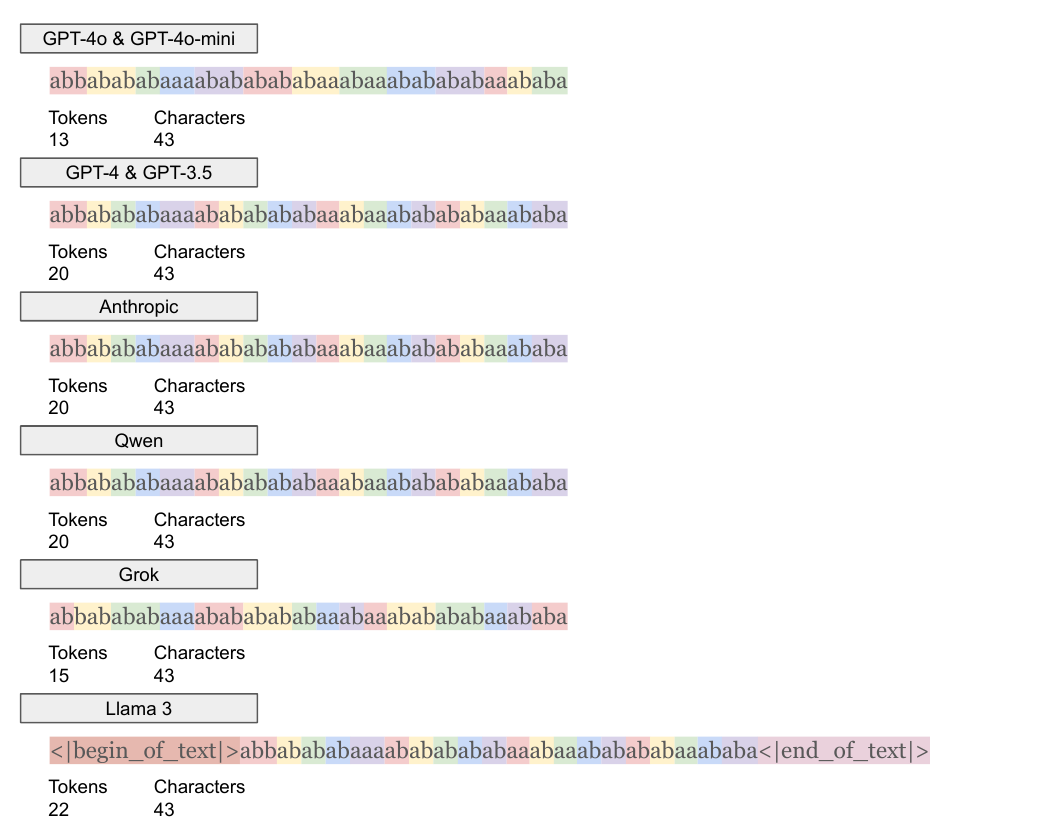}
    \vspace{-0.5em}
    \caption{Difference in tokenization on \textit{long} \textit{binary} strings without punctuations across different LLMs.}
    \label{fig:tokenlongall}
    \vspace{-1.5em}
\end{figure*}

\begin{figure*}[htbp]
    \centering
    \includegraphics[width=\linewidth]{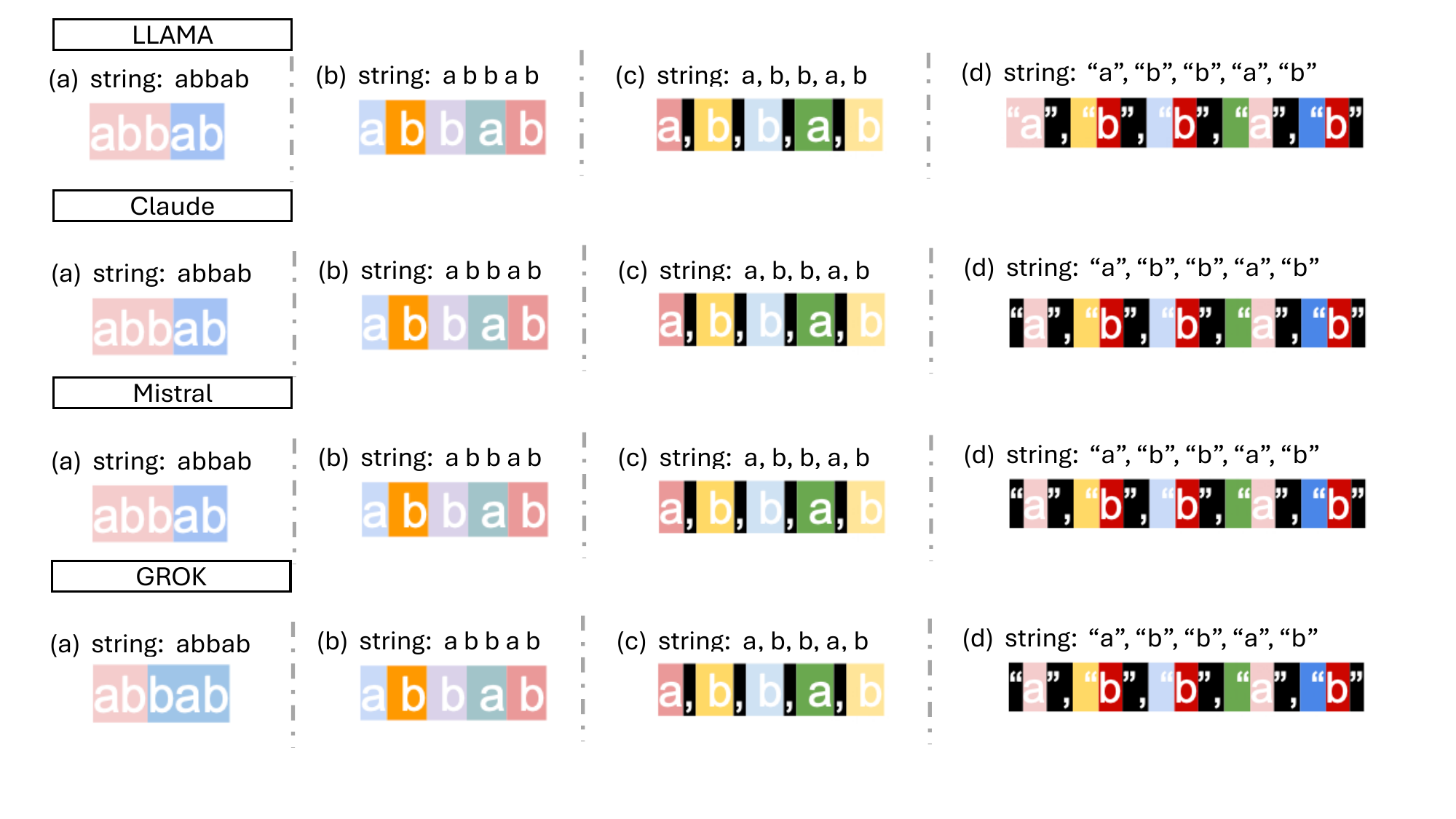}
    \vspace{-0.5em}
    \caption{Difference in tokenization on \textit{binary} strings when counting instances are presented in different formats with punctuations to facilitate tokenization, across different LLMs. }
    \label{fig:tokenall}
    \vspace{-1.5em}
\end{figure*}

Figure~\ref{fig:tokenlongall} and Figure~\ref{fig:tokenall} illustrate the tokenization of input binary strings with difference lengths across various LLMs. We investigate both language models and multi-modal models, observing nearly identical tokenization behaviors across most tested models (except GPT-4o series). Therefore, in Figure~\ref{fig:reverse_token} and Figure~\ref{fig:sort_token}, we use the GPT-4o series models to further demonstrate tokenization patterns in more complex cases. These include random letter strings, random number sequences, dictionary words, high-frequency words, and mixed digit-letter strings. This analysis provides insight into how tokenization varies across different input types and structural formats.

 Since pure strings may be tokenized differently due to the varying byte pair encoding (BPE) schemes used by each tokenizer. When a token-level delimiter is introduced, we consistently observe that the delimiter is combined with the adjacent letter, aligning with our previous assumptions. Additionally, some models handle the initial token differently, resulting in the first letter being treated as a standalone token (e.g., in the Grok model) or being combined with a quotation delimiter (e.g., in GPT-4o mini).
In summary, our string design effectively allows us to manipulate modern LLMs to tokenize identical counting instances into different, desired tokens.
\label{app:token}

\section{Prompt Template}
\label{app: prompttemplate}
The set of prompts we use for counting experiments (base, unsupervised CoT, supervised CoT) are shown in Figure~\ref{fig:counting_prompts_all}.
The set of prompts we use for reversing and sorting experiments (unsupervised CoT, supervised CoT) are shown in Figure~\ref{fig:reversing_prompts_all} and Figure~\ref{fig:sorting_prompts_all} respectively.

\section{Case Studies: Counting}
\label{app:cs_count}
In this section, we use counting task to showcase our experiment results. The tables present cases for each type of token
counted using CoT. As shown, Supervised CoT consistently adheres to a strict step-by-step template, accurately extracting the counter at each step. In contrast, Unsupervised CoT often skips crucial steps or deviates from the optimal method of extracting counters. Additionally, Supervised CoT with type (d) tokenization tends to produce much longer reasoning contexts, yet still achieves the best performance due to the combination of optimal tokenization and supervision.

We also repeated the experiments with Claude 3.5 Sonnet and Qwen Turbo. 
With Qwen model, we observed similar trends as GPT-4o mini (Table ~\ref{tab:qwen}). With Claude model, there is a slight exception that type (c) yielded the best results among types (a)-(d), as shown in Table~\ref{tab:claude}. Upon investigation, we suspect this is because type (d) results in longer CoT steps due to the higher number of irrelevant tokens generation, leading to long-context reasoning failures in many cases for this model.
We also provide case studies using GPT-4o mini for counting tasks, including examples where CoT led to both correct and incorrect answers. Additionally, we reveal the inferior performance with OpenAI o1 full model when tokenization is not properly done, detailed in Appendix section \ref{appendix:o1}.

To this end, we are confident that our experimental results can be generalized to other LLMs, given that the training methods and tokenization strategies (as demonstrated in Appendix Section \ref{app:token}) are nearly identical, leading to counting being performed in a similar manner across such models.

\subsection{OpenAI o1}
\label{appendix:o1}
We evaluate o1 on samples ranging from 30 to 40 letters in length, using pure string (type (a)) to showcase the importance of using proper tokenization. Additionally, since o1 applies inference-time scaling techniques (such as MCTS search and per-step verifying), it implicitly engages in advanced chain-of-thought reasoning. Thus, we rely solely on a straightforward prompt rather than explicitly specifying reasoning steps using Supervised Chain-of-Thought as with other naive models. The final accuray on tested samples for o1 is 50\%, which is much lower than using GPT-4o mini with most optimal tokenization techniques (Table \ref{tab:main_result_gpt}, ~70\% in such length range). An example for correct counting is shown in Table \ref{tab:casestudyo1incorrect}, and an example for incorrect counting is shown in Table \ref{tab:casestudyo1correct}. \textbf{In conclusion, advanced LLM searching algorithms and inference time scaling techniques do not make up for defect in tokenizer. }

\subsection{Qwen Turbo}
We evaluate Qwen Turbo using supervised Chain-of-Thought (CoT) prompts.
Tables ~\ref{tab:casestudyqwenincorrect} and ~\ref{tab:casestudyqwenincorrectcont} demonstrate an incorrect counting example using tokenization type (a),
Table ~\ref{tab:casestudyqwencorrect} shows a correct counting example using tokenization type (d).

\begin{table}[t]
\centering
\resizebox{0.47\textwidth}{!}{
\begin{tabular}{c|cccccc}
\toprule
\multirow{1}{*}{\textbf{string-token }} & \multicolumn{2}{c}{\textbf{len $\in$ [10-20]}} & \multicolumn{2}{c}{\textbf{len $\in$ [20-30]}} & \multicolumn{2}{c}{\textbf{len $\in$ [30-40]}} \\
\cmidrule(lr){2-3} \cmidrule(lr){4-5} \cmidrule(lr){6-7}
 \textbf{type} & \textbf{count \texttt{a}} & \textbf{count \texttt{b}} & \textbf{count \texttt{a}} & \textbf{count \texttt{b}} & \textbf{count \texttt{a}} & \textbf{count \texttt{b}} \\
\midrule
\textbf{(a)} & 56.40 & 62.50 & \colorcellred{26.20}{0}{40}26.20 & \colorcellred{32.20}{0}{40}32.20 & \colorcellred{16.20}{0}{40}16.20 & \colorcellred{15.90}{0}{40}15.90 \\
\textbf{(b)} & \colorcellgreen{75.10}{70}{100}75.10 & \colorcellgreen{80}{70}{100}80.00 & \colorcellgreen{80.40}{80}{100}50.60& 54.60& \colorcellred{31.10}{0}{40}31.10 & \colorcellred{28.30}{0}{40}28.30\\
\textbf{(c)} & \colorcellgreen{93.40}{70}{100}93.40 & \colorcellgreen{96.00}{70}{100}96.00  & \colorcellgreen{81.60}{70}{100}81.60 & \colorcellgreen{83.50}{70}{100}83.50 &59.20 &57.60\\
\textbf{(d)} & \colorcellgreen{95.90}{70}{100}95.90 & \colorcellgreen{96.70}{70}{100}96.70 &\colorcellgreen{83.20}{70}{100}83.20& \colorcellgreen{86.50}{70}{100}86.50& 68.10& 63.50 \\
\bottomrule
\end{tabular}

}
\caption{Counting results under the same settings as in Table \ref{tab:claude} but using Qwen Turbo API. Numbers indicate the average accuracy (\%) over 1000 random generated instances.}
\label{tab:qwen}
\end{table}
Notably, Qwen Turbo generates more tokens per CoT step compared to Claude 3.5 Sonnet, which appears to lead to its slightly lower performance. This suggests that concise reasoning steps is beneficial for counting accuracy.

\subsection{GPT-4o mini}
\label{app:gptcasestudy}
We present three progressive examples that demonstrate the effectiveness of combining Chain-of-Thought (CoT) reasoning with appropriate tokenization strategies:
\begin{enumerate}
    \item Table ~\ref{tab:casestudygpt4ominiincorrect1} shows that using a base prompt with standard BPE tokenization (type (a)) results in a counting error of 3 from the correct value.
    \item Table ~\ref{tab:casestudygpt4ominiincorrect2} demonstrates that incorporating supervised CoT improves accuracy, reducing the counting error to just 1.
    \item Table ~\ref{tab:casestudygpt4ominicorrect} illustrates that combining supervised CoT with type (d) character-wise tokenization achieves perfect accuracy, matching the gold label exactly.
\end{enumerate}
These examples clearly demonstrate how the synergy between CoT reasoning and appropriate tokenization methods can progressively enhance counting accuracy.

\begin{figure*}[t]
    \centering
    \includegraphics[width=\linewidth]{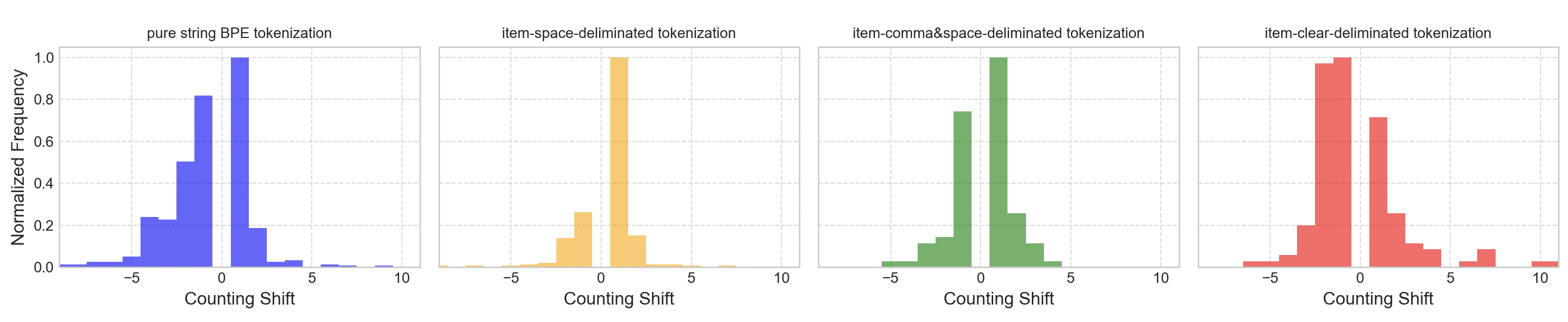}
    
    \caption{Same error-shifting distribution (as in Figure \ref{fig:shift}) but for Claude model. Claude 3.5 tend to count \textit{more} than counting \textit{less}, compared to GPT-4o. }
    \label{fig:shiftclaude}

\end{figure*}
\subsection{Claude 3.5 Sonnet}
We evaluate Claude 3.5 Sonnet using supervised CoT prompts. We provide example cases demonstrating different tokenization approaches and their outcomes:
\begin{itemize}
    \item Tables ~\ref{tab:casestudyclaudeincorrect} and ~\ref{tab:casestudyclaudeincorrectcont} showcase incorrect counting results using tokenization type (a)
    \item Tables ~\ref{tab:casestudyclaudecorrect} and ~\ref{tab:casestudyclaudecorrectcont} demonstrate correct counting using tokenization type (d)
    \item Tables ~\ref{tab:casestudyclaudecorrect2} and ~\ref{tab:casestudyclaudecorrect2cont} illustrate correct counting using tokenization type (c)
\end{itemize}

To sum up, our analyses show that tokenization type (c) yields superior results compared to type (d).
Notably, in Tables ~\ref{tab:casestudyclaudecorrect} and ~\ref{tab:casestudyclaudecorrectcont}, we observe that type (d) tokenization generates excessive and irrelevant content (specifically, index information) which may interfere with the accuracy of the counting process.

\section{Replication Experiments Note}
We have open-sourced the experimental results for every instance of each experiment, in the provided GitHub link, to facilitate future research and analysis by other researchers. All reported experiment numbers are stable, using the same experimental settings and prompts. Specifically, we observe an average variance in accuracy of less than 1\% across runs of the same experiments, indicating that they are fully replicable with the same model version used. Note that updates to the API version may cause potential variations in results, which are beyond our control.

\newpage

\begin{figure*}[htbp]
    \centering
    \begin{subfigure}{0.95\textwidth} 
        \begin{promptbox}{Base Prompt}
            Count the number of appearances of `\{substring\}'s in the string below. Directly output `Result: ' followed by the counted number. Do not use bold font in the response.
            
            \vspace{2mm}

            String: \{sample\}
            
            \vspace{2mm}
            
            Response:
        \end{promptbox}
        \caption{Base prompt template}
        \label{fig:count_base_prompt}
    \end{subfigure}
    
    \vspace{4mm}
    
    \begin{subfigure}{0.95\textwidth} 
        \begin{promptbox}{Chain-of-Thought (CoT) Prompt}
            Determine the number of appearances of `\{substring\}'s in the string below. Think step by step. Directly output `Result: ' followed by the counted number. Do not use bold font in the response.

            \vspace{2mm}
            
            String: \{sample\}
            
            \vspace{2mm}
            
            Response:
        \end{promptbox}
        \caption{Chain-of-Thought (unsupervised) prompt template}
        \label{fig:count_cot_prompt}
    \end{subfigure}
    
    \vspace{4mm}
    
    \begin{subfigure}{0.95\textwidth} 
        \begin{promptbox}{Supervised Chain-of-Thought Prompt}
            Task: Count the number of occurrences of the substring `\{substring\}' in the given string step by step. Do not use bold font in the response.
            
            \vspace{2mm}

            Instructions:\\
            • Iterate through the string and count each occurrence of the substring.\\
            • Use a counter to keep track of how many times the substring appears.\\
            • Output your step-by-step counting process.\\
            • Conclude with: `Result: ' followed by the number.
            
            \vspace{2mm}
            
            String: \{sample\}
            
            \vspace{2mm}
            
            Response:
        \end{promptbox}
        \caption{Chain-of-Thought (supervised) prompt template}
        \label{fig:count_scot_prompt}
    \end{subfigure}
    
    \caption{Different prompt templates used in counting experiments. The templates include a base prompt, an unsupervised chain-of-thought prompt, and a supervised chain-of-thought prompt.}
    \label{fig:counting_prompts_all}
\end{figure*}
\newpage

\begin{figure*}[htbp]
    \centering
    \begin{subfigure}{0.95\textwidth} 
        \begin{promptbox}{Chain-of-Thought (CoT) Prompt}
            Reverse the string provided below. Think step  by step. Output the final result in a dictionary with the key `Result'. For instance, if the input string is `iamhappy', the output would be: \{`Result': `yppahmai'\}. The input string may contain punctuations, do not add them into reversed string.

            \vspace{2mm}
            
            Input string: \{\{string\}\}
            
        \end{promptbox}
        \caption{Chain-of-Thought (unsupervised) prompt template}
        \label{fig:reverse_cot_prompt}
    \end{subfigure}
    
    \vspace{4mm}
    
    \begin{subfigure}{0.95\textwidth} 
        \begin{promptbox}{Supervised Chain-of-Thought Prompt}
            Reverse the string following the instructions below:
            \begin{enumerate}
                \item Initialize an empty string called `reversed'
                \item For each character in the input string:
                \begin{itemize}
                    \item Remove the first (leftmost) character
                    \item Add this character to the beginning of `reversed'
                    \item Display the remaining input string
                \end{itemize}
                \item Once all characters are processed, `reversed' will contain the reversed string. Return the reversed string in a dictionary format with the key `Result'. For example, the result dictionary for input string "hello" will be \{`Result': `olleh'\}.
            \end{enumerate}

            \vspace{2mm}
            
            Input String: \{\{string\}\}
            
        \end{promptbox}
        \caption{Chain-of-Thought (supervised) prompt template}
        \label{fig:reverse_scot_prompt}
    \end{subfigure}
    
    \caption{Different prompt templates used in reversing experiments. The templates include an unsupervised chain-of-thought prompt, and a supervised chain-of-thought prompt.}
    \label{fig:reversing_prompts_all}
\end{figure*}

\newpage
\begin{figure*}[htbp]
    \centering
    \begin{subfigure}{0.95\textwidth} 
        \begin{promptbox}{Chain-of-Thought (CoT) Prompt}
            Sort the string below in ascending order using insertion sort. Think step by step. Note that lower case characters are greater than upper case characters. Convert the sorted characters into a single string, and store the result in a dictionary with the key `Result.' For example, given the string `2CsU4bSc', the sorted result should be `24CSUbcs' and should be stored as \{`Result': `24CSUbcs'\}.

            \vspace{2mm}
            
            String: \{\{string\}\}
            
        \end{promptbox}
        \caption{Chain-of-Thought (unsupervised) prompt template}
        \label{fig:sort_cot_prompt}
    \end{subfigure}
    
    \vspace{4mm}
    
    \begin{subfigure}{0.95\textwidth} 
        \begin{promptbox}{Supervised Chain-of-Thought Prompt}
            Sort the string below in ascending order using insertion sort following the steps below. Note that lower case characters are greater than upper case characters.
            \begin{enumerate}
                \item Start by creating an empty list for sorted characters.
                \item For each step:
                \begin{itemize}
                    \item Specify which character you're currently looking and insert the character to the correct place in sorted list.
                    \item Display both the original string (showing characters remaining) and sorted list after each step.
                \end{itemize}
                \item Conbine the sorted list to a single string. Return the sorted string in a dictionary format with the key `Result'. For example, the result dictionary for input list ``[`2', `C', `s', `U', `4', `b', `S', `c']'' will be \{`Result': `24CSUbcs'\}.
            \end{enumerate}

            \vspace{2mm}
            String: \{\{string\}\}
        \end{promptbox}
        \caption{Chain-of-Thought (supervised) prompt template}
        \label{fig:sort_scot_prompt}
    \end{subfigure}
    
    \caption{Different prompt templates used in sorting experiments. The templates include an unsupervised chain-of-thought prompt, and a supervised chain-of-thought prompt.}
    \label{fig:sorting_prompts_all}
\end{figure*}
\newpage

\begin{table*}[htbp]
    \centering
    \small
    \begin{tcolorbox}[
        colback=white,
        boxrule=1.2pt,
        arc=8pt,
        outer arc=8pt,
        width=0.95\textwidth
    ]
    \begin{tabular}{p{0.93\textwidth}}
     \rowcolor{gray!10} \multicolumn{1}{p{\linewidth}}{\textbf{Model \& Experiment:}} \\[0.4em]
    OpenAI o1, token type (a), length 30-40 \\[0.4em]
    \midrule[1.2pt]
    \rowcolor{gray!10} \multicolumn{1}{p{\linewidth}}{\textbf{Prompt:}} \\[0.4em]
    \begin{minipage}{0.93\textwidth}
Count the number of appearances of `a' in the string below. Do not use tool.\\\\String: abaaaaaaabaabaaabbaaaaabaabbbbbbbabaaaaa\\\\Response:
    
    \end{minipage} \\[0.8em]
    \midrule[1.2pt]
    
    \rowcolor{gray!10} \multicolumn{1}{p{\linewidth}}{\textbf{Model Response:}} \\[0.4em]
    \begin{minipage}{0.93\textwidth}
     Let's count the occurrences of `a' in the given string:\\\\    String:  \\    \texttt{\\    abaaaaaaabaabaaabbaaaaabaabbbbbbbabaaaaa\\}   \\  We'll go through each character and count how many times `a' appears:\\    \texttt{ \\    a (1)\\    b\\    a (2)\\    a (3)\\    a (4)\\    a (5)\\    a (6)\\    a (7)\\    a (8)\\    b\\    a (9)\\   a (10)\\    b\\   a (11)\\    a (12)\\    a (13)\\    b\\    b\\    a (14)\\    a (15)\\    a (16)\\    a (17)\\    a (18)\\    b\\    a (19)\\    a (20)\\    b\\    b\\    b\\    b\\    b\\    b\\    b\\    a (21)\\    b\\    a (22)\\    a (23)\\    a (24)\\    a (25)\\    a (26)\\    \textcolor{red}{a (27)}\\    }\\    Number of `a's counted: **\textcolor{red}{27}**    
    \end{minipage} \\[0.8em]
    \midrule[1.2pt]
    
    \rowcolor{gray!10} \multicolumn{1}{p{\linewidth}}{\textbf{Correct Answer:}} \\[0.4em]
     26\\[0.4em]
    
    \end{tabular}
    \end{tcolorbox}
    \caption{Case Study 1 - Incorrect Prediction with o1 (base prompt)}
    \label{tab:casestudyo1incorrect}
\end{table*}
\begin{table*}[htbp]
    \centering
    \small
    \begin{tcolorbox}[
        colback=white,
        boxrule=1.2pt,
        arc=8pt,
        outer arc=8pt,
        width=0.95\textwidth
    ]
    \begin{tabular}{p{0.93\textwidth}}
     \rowcolor{gray!10} \multicolumn{1}{p{\linewidth}}{\textbf{Model \& Experiment:}} \\[0.4em]
    OpenAI o1, token type (a), length 30-40 \\[0.4em]
    \midrule[1.2pt]
    \rowcolor{gray!10} \multicolumn{1}{p{\linewidth}}{\textbf{Prompt:}} \\[0.4em]
    \begin{minipage}{0.93\textwidth}
Count the number of appearances of `b' in the string below. Do not use tool.\\\\String: babbbbbbbabbababbbaabbbbbbabaabbbbabb\\\\Response:
    
    \end{minipage} \\[0.8em]
    \midrule[1.2pt]
    
    \rowcolor{gray!10} \multicolumn{1}{p{\linewidth}}{\textbf{Model Response:}} \\[0.4em]
    \begin{minipage}{0.93\textwidth}
       Let's count carefully:\\\\   The given string is:  \\    \texttt{babbbbbbbabbababbbaabbbbbbabaabbbbabb}\\\\    Let's go through it character by character and count every `b':\\\\    \texttt{\\    b (1)  \\    a  \\    b (2)  \\    b (3)  \\    b (4)  \\    b (5)  \\    b (6)  \\    b (7)  \\    b (8)  \\    a  \\    b (9)  \\    b (10)  \\    a  \\    b (11)  \\    a  \\    b (12)  \\    b (13)  \\    b (14)  \\    a  \\    a  \\    b (15)  \\    b (16)  \\    b (17)  \\    b (18)  \\    b (19)  \\    b (20)  \\    a  \\    b (21)  \\    a  \\    a  \\    b (22)  \\    b (23)  \\    b (24)  \\    b (25)  \\    a  \\    b (26)  \\    b (27)\\    }\\\\    Counting all the marked `b's, we get a total of **\textcolor{teal}{27}**.
    \end{minipage} \\[0.8em]
    \midrule[1.2pt]
    
    \rowcolor{gray!10} \multicolumn{1}{p{\linewidth}}{\textbf{Correct Answer:}} \\[0.4em]
    27 \\[0.4em]
    
    \end{tabular}
    \end{tcolorbox}
    \caption{Case Study 2 - Correct Prediction with o1 (base prompt)}
    \label{tab:casestudyo1correct}
\end{table*}
\newpage

\begin{table*}[htbp]
    \centering
    \small
    \begin{tcolorbox}[
        colback=white,
        boxrule=1.2pt,
        arc=8pt,
        outer arc=8pt,
        width=0.95\textwidth
    ]
    \begin{tabular}{p{0.93\textwidth}}
     \rowcolor{gray!10} \multicolumn{1}{p{\linewidth}}{\textbf{Model \& Experiment:}} \\[0.4em]
    Claude 3.5 Sonnet, token type (a), length 30-40 \\[0.4em]
    \midrule[1.2pt]
    \rowcolor{gray!10} \multicolumn{1}{p{\linewidth}}{\textbf{Prompt:}} \\[0.4em]
    \begin{minipage}{0.93\textwidth}
Task: Count the number of occurrences of the substring `a' in the given string step by step. Do not use bold font in the response.\\

    Instructions:
    \begin{itemize}
        \item Iterate through the string and count each occurrence of the substring.
        \item Use a counter to keep track of how many times the substring appears.
        \item Output your step-by-step counting process.
        \item Conclude with: `Result: ' followed by the number.
    \end{itemize}

    String: babbbababbbaabbbbaabaabaaabbaaababbbabb\\\\
    Response:
    \end{minipage} \\[0.8em]
    \midrule[1.2pt]
     \rowcolor{gray!10} \multicolumn{1}{p{\linewidth}}{\textbf{Model Response:}} \\[0.4em]
    \begin{minipage}{0.93\textwidth}
       Step 1: Initialize a counter to 0.\\Counter = 0\\\\Step 2: Start iterating through the string:\\b - not `a', counter remains 0\\a - found `a', increment counter\\Counter = 1\\\\Step 3: Continue iterating:\\b - not `a', counter remains 1\\b - not `a', counter remains 1\\b - not `a', counter remains 1\\b - not `a', counter remains 1\\a - found `a', increment counter\\Counter = 2\\\\Step 4: Continue iterating:\\b - not `a', counter remains 2\\a - found `a', increment counter\\Counter = 3\\\\Step 5: Continue iterating:\\b - not `a', counter remains 3\\b - not `a', counter remains 3\\b - not `a', counter remains 3\\a - found `a', increment counter\\Counter = 4\\\\Step 6: Continue iterating:\\a - found `a', increment counter\\Counter = 5\\\\Step 7: Continue iterating:\\b - not `a', counter remains 5\\b - not `a', counter remains 5\\b - not `a', counter remains 5\\b - not `a', counter remains 5\\a - found `a', increment counter\\Counter = 6\\\\Step 8: Continue iterating:\\a - found `a', increment counter\\Counter = 7
       \end{minipage}\\[0.8em]
    
    \end{tabular}
    \end{tcolorbox}
    \caption{Case Study 3 - Incorrect Prediction with Claude 3.5 Sonnet (CoT prompt)}
    \label{tab:casestudyclaudeincorrect}
\end{table*}

\begin{table*}[htbp]
    \centering
    \small
    \begin{tcolorbox}[
        colback=white,
        boxrule=1.2pt,
        arc=8pt,
        outer arc=8pt,
        width=0.95\textwidth
    ]
    \begin{tabular}{p{0.93\textwidth}}
    \begin{minipage}{0.93\textwidth}
    Step 9: Continue iterating:\\b - not `a', counter remains 7\\a - found `a', increment counter\\Counter = 8\\\\
       Step 10: Continue iterating:\\a - found `a', increment counter\\Counter = 9\\\\Step 11: Continue iterating:\\b - not `a', counter remains 9\\a - found `a', increment counter\\Counter = 10\\\\Step 12: Continue iterating:\\a - found `a', increment counter\\Counter = 11\\\\Step 13: Continue iterating:\\b - not `a', counter remains 11\\b - not `a', counter remains 11\\a - found `a', increment counter\\Counter = 12\\\\Step 14: Continue iterating:\\a - found `a', increment counter\\Counter = 13\\\\Step 15: Finish iterating through the remaining characters:\\b - not `a', counter remains 13\\a - found `a', increment counter\\Counter = 14\\\\Step 16: Iteration complete.\\\\Result: \textcolor{red}{14}
       \end{minipage}\\[0.8em]
    \midrule[1.2pt]
    
    \rowcolor{gray!10} \multicolumn{1}{p{\linewidth}}{\textbf{Correct Answer:}} \\[0.4em]
     17\\[0.4em]
    \end{tabular}
    \end{tcolorbox}
    \caption{Case Study 3 Continued}
    \label{tab:casestudyclaudeincorrectcont}
\end{table*}

\begin{table*}[htbp]
    \centering
    \small
    \begin{tcolorbox}[
        colback=white,
        boxrule=1.2pt,
        arc=8pt,
        outer arc=8pt,
        width=0.95\textwidth
    ]
    \begin{tabular}{p{0.93\textwidth}}
     \rowcolor{gray!10} \multicolumn{1}{p{\linewidth}}{\textbf{Model \& Experiment:}} \\[0.4em]
    Claude 3.5 Sonnet, token type (d), length 30-40 \\[0.4em]
    \midrule[1.2pt]
    \rowcolor{gray!10} \multicolumn{1}{p{\linewidth}}{\textbf{Prompt:}} \\[0.4em]
    \begin{minipage}{0.93\textwidth}
Task: Count the number of occurrences of the substring 'a' in the given string step by step. Do not use bold font in the response.\\\\
    Instructions:
    \begin{itemize}
        \item Iterate through the string and count each occurrence of the substring.
        \item Use a counter to keep track of how many times the substring appears.
        \item Output your step-by-step counting process.
        \item Conclude with: `Result: ' followed by the number.
    \end{itemize}

    String: [`b', `a', `b', `b', `b', `a', `b', `a', `b', `b', `b', `a', `a', `b', `b', `b', `b', `a', `a', `b', `a', `a', `b', `a', `a', `a', `b', `b', `a', `a', `a', `b', `a', `b', `b', `b', `a', `b', `b']\\\\
    Response:
    \end{minipage} \\[0.8em]
    \midrule[1.2pt]
     \rowcolor{gray!10} \multicolumn{1}{p{\linewidth}}{\textbf{Model Response:}} \\[0.4em]
    \begin{minipage}{0.93\textwidth}
    Step 1: Initialize counter to 0\\Counter: 0\\\\Step 2: Start iterating through the string
    
    [`b', `a', `b', `b', `b', `a', `b', `a', `b', `b', `b', `a', `a', `b', `b', `b', `b', `a', `a', `b', `a', `a', `b', `a', `a', `a', `b', `b', `a', `a', `a', `b', `a', `b', `b', `b', `a', `b', `b']\\\\Step 3: Found `a' at index 1\\Counter: 1\\\\Step 4: Found `a' at index 5\\Counter: 2\\\\Step 5: Found `a' at index 7\\Counter: 3\\\\Step 6: Found `a' at index 11\\Counter: 4\\\\Step 7: Found `a' at index 12\\Counter: 5\\\\Step 8: Found `a' at index 17\\Counter: 6\\\\Step 9: Found `a' at index 18\\Counter: 7\\\\Step 10: Found `a' at index 20\\Counter: 8\\\\Step 11: Found `a' at index 21\\Counter: 9\\\\
       \end{minipage}\\[0.8em]
    \end{tabular}
    \end{tcolorbox}
    \caption{Case Study 4 - Correct Prediction with Claude 3.5 Sonnet 1 (CoT prompt)}
    \label{tab:casestudyclaudecorrect}
\end{table*}

\begin{table*}[htbp]
    \centering
    \small
    \begin{tcolorbox}[
        colback=white,
        boxrule=1.2pt,
        arc=8pt,
        outer arc=8pt,
        width=0.95\textwidth
    ]
    \begin{tabular}{p{0.93\textwidth}}
    \begin{minipage}{0.93\textwidth}
       Step 12: Found `a' at index 23\\Counter: 10\\\\Step 13: Found `a' at index 24\\Counter: 11\\\\Step 14: Found `a' at index 25\\Counter: 12\\\\Step 15: Found `a' at index 28\\Counter: 13\\\\Step 16: Found `a' at index 29\\Counter: 14\\\\Step 17: Found `a' at index 30\\Counter: 15\\\\Step 18: Found `a' at index 32\\Counter: 16\\\\Step 19: Found `a' at index 36\\Counter: 17\\\\Step 20: Finished iterating through the string\\\\Result: \textcolor{teal}{17}
       \end{minipage}\\[0.8em]
    \midrule[1.2pt]
    
    \rowcolor{gray!10} \multicolumn{1}{p{\linewidth}}{\textbf{Correct Answer:}} \\[0.4em]
     17\\[0.4em]
    \end{tabular}
    \end{tcolorbox}
    \caption{Case Study 4 Continued}
    \label{tab:casestudyclaudecorrectcont}
\end{table*}

\begin{table*}[htbp]
    \centering
    \small
    \begin{tcolorbox}[
        colback=white,
        boxrule=1.2pt,
        arc=8pt,
        outer arc=8pt,
        width=0.95\textwidth
    ]
    \begin{tabular}{p{0.93\textwidth}}
     \rowcolor{gray!10} \multicolumn{1}{p{\linewidth}}{\textbf{Model \& Experiment:}} \\[0.4em]
    Claude 3.5 Sonnet, token type (c), length 30-40 \\[0.4em]
    \midrule[1.2pt]
    \rowcolor{gray!10} \multicolumn{1}{p{\linewidth}}{\textbf{Prompt:}} \\[0.4em]
    \begin{minipage}{0.93\textwidth}
Task: Count the number of occurrences of the substring 'a' in the given string step by step. Do not use bold font in the response.\\\\
    Instructions:
    \begin{itemize}
        \item Iterate through the string and count each occurrence of the substring.
        \item Use a counter to keep track of how many times the substring appears.
        \item Output your step-by-step counting process.
        \item Conclude with: `Result: ' followed by the number.
    \end{itemize}

    String: b, a, b, b, b, a, b, a, b, b, b, a, a, b, b, b, b, a, a, b, a, a, b, a, a, a, b, b, a, a, a, b, a, b, b, b, a, b, b\\\\
    Response:
    \end{minipage} \\[0.8em]
    \midrule[1.2pt]
     \rowcolor{gray!10} \multicolumn{1}{p{\linewidth}}{\textbf{Model Response:}} \\[0.4em]
    \begin{minipage}{0.93\textwidth}
    Step 1: Initialize counter to 0\\Counter: 0\\\\Step 2: Start iterating through the string\\b, - Counter: 0\\a, - Found `a'. Counter: 1\\b, - Counter: 1\\b, - Counter: 1\\b, - Counter: 1\\a, - Found `a'. Counter: 2\\b, - Counter: 2\\a, - Found `a'. Counter: 3\\b, - Counter: 3\\b, - Counter: 3\\b, - Counter: 3\\a, - Found `a'. Counter: 4\\a, - Found `a'. Counter: 5\\b, - Counter: 5\\b, - Counter: 5\\b, - Counter: 5\\b, - Counter: 5\\a, - Found 'a'. Counter: 6\\a, - Found 'a'. Counter: 7\\b, - Counter: 7\\a, - Found 'a'. 
       \end{minipage}\\[0.8em]
    \end{tabular}
    \end{tcolorbox}
    \caption{Case Study 5 - Correct Prediction with Claude 3.5 Sonnet 2 (CoT prompt)}
    \label{tab:casestudyclaudecorrect2}
\end{table*}

\begin{table*}[htbp]
    \centering
    \small
    \begin{tcolorbox}[
        colback=white,
        boxrule=1.2pt,
        arc=8pt,
        outer arc=8pt,
        width=0.95\textwidth
    ]
    \begin{tabular}{p{0.93\textwidth}}
    \begin{minipage}{0.93\textwidth}
       Counter: 8\\a, - Found 'a'. Counter: 9\\b, - Counter: 9\\a, - Found 'a'. Counter: 10\\a, - Found 'a'. Counter: 11\\a, - Found 'a'. Counter: 12\\b, - Counter: 12\\b, - Counter: 12\\a, - Found `a'. Counter: 13\\a, - Found 'a'. Counter: 14\\a, - Found `a'. Counter: 15\\b, - Counter: 15\\a, - Found `a'. Counter: 16\\b, - Counter: 16\\b, - Counter: 16\\b, - Counter: 16\\a, - Found `a'. Counter: 17\\b, - Counter: 17\\b - Counter: 17\\\\Step 3: Finished iterating through the string\\\\Result: \textcolor{teal}{17}
       \end{minipage}\\[0.8em]
    \midrule[1.2pt]
    
    \rowcolor{gray!10} \multicolumn{1}{p{\linewidth}}{\textbf{Correct Answer:}} \\[0.4em]
     17\\[0.4em]
    \end{tabular}
    \end{tcolorbox}
    \caption{Case Study 5 Continued}
    \label{tab:casestudyclaudecorrect2cont}
\end{table*}

\begin{table*}[htbp]
    \centering
    \small
    \begin{tcolorbox}[
        colback=white,
        boxrule=1.2pt,
        arc=8pt,
        outer arc=8pt,
        width=0.95\textwidth
    ]
    \begin{tabular}{p{0.93\textwidth}}
     \rowcolor{gray!10} \multicolumn{1}{p{\linewidth}}{\textbf{Model \& Experiment:}} \\[0.4em]
    Qwen Turbo, token type (a), length 30-40 \\[0.4em]
    \midrule[1.2pt]
    \rowcolor{gray!10} \multicolumn{1}{p{\linewidth}}{\textbf{Prompt:}} \\[0.4em]
    \begin{minipage}{0.93\textwidth}
Task: Count the number of occurrences of the substring `a' in the given string step by step. Do not use bold font in the response.\\

    Instructions:
    \begin{itemize}
        \item Iterate through the string and count each occurrence of the substring.
        \item Use a counter to keep track of how many times the substring appears.
        \item Output your step-by-step counting process.
        \item Conclude with: `Result: ' followed by the number.
    \end{itemize}

    String: baaabaaabaaababbbababbaababbabbaabb\\\\
    Response:
    \end{minipage} \\[0.8em]
    \midrule[1.2pt]
     \rowcolor{gray!10} \multicolumn{1}{p{\linewidth}}{\textbf{Model Response:}} \\[0.4em]
    \begin{minipage}{0.93\textwidth}
    To count the number of occurrences of the substring `a' in the given string step by step, we will iterate through the string and increment a counter each time we encounter an `a'. \\\\String: baaabaaabaaababbbababbaababbabbaabb\\\\Step 1: Start with the first character `b', no `a' found.\\Step 2: Second character is `a', increment the counter (counter = 1).\\Step 3: Third character is `a', increment the counter (counter = 2).\\Step 4: Fourth character is `a', increment the counter (counter = 3).\\Step 5: Fifth character is `b', no `a' found.\\Step 6: Sixth character is `a', increment the counter (counter = 4).\\Step 7: Seventh character is `a', increment the counter (counter = 5).\\Step 8: Eighth character is `b', no `a' found.\\Step 9: Ninth character is `a', increment the counter (counter = 6).\\Step 10: Tenth character is `a', increment the counter (counter = 7).\\Step 11: Eleventh character is `b', no `a' found.\\Step 12: Twelfth character is `a', increment the counter (counter = 8).\\Step 13: Thirteenth character is `b', no `a' found.\\Step 14: Fourteenth character is `b', no `a' found.\\Step 15: Fifteenth character is `b', no `a' found.\\Step 16: Sixteenth character is `a', increment the counter (counter = 9).\\Step 17: Seventeenth character is `b', no `a' found.\\Step 18: Eighteenth character is `a', increment the counter (counter = 10).\\
       \end{minipage}\\[0.8em]
    
    \end{tabular}
    \end{tcolorbox}
    \caption{Case Study 6 - Incorrect Prediction with Qwen Turbo (CoT prompt)}
    \label{tab:casestudyqwenincorrect}
    
\end{table*}

\begin{table*}[htbp]
    \centering
    \small
    \begin{tcolorbox}[
        colback=white,
        boxrule=1.2pt,
        arc=8pt,
        outer arc=8pt,
        width=0.95\textwidth
    ]
    \begin{tabular}{p{0.93\textwidth}}
    \begin{minipage}{0.93\textwidth}
    Step 19: Nineteenth character is `b', no 'a' found.\\Step 20: Twentieth character is `b', no `a' found.\\Step 21: Twenty-first character is `a', increment the counter (counter = 11).\\Step 22: Twenty-second character is `a', increment the counter (counter = 12).\\Step 23: Twenty-third character is `b', no 'a' found.\\Step 24: Twenty-fourth character is `b', no `a' found.\\Step 25: Twenty-fifth character is `a', increment the counter (counter = 13).\\Step 26: Twenty-sixth character is `b', no `a' found.\\Step 27: Twenty-seventh character is `a', increment the counter (counter = 14).\\Step 28: Twenty-eighth character is `b', no `a' found.\\Step 29: Twenty-ninth character is `b', no `a' found.\\Step 30: Thirtieth character is `a', increment the counter (counter = 15).\\Step 31: Thirty-first character is `b', no `a' found.\\Step 32: Thirty-second character is `b', no `a' found.\\Step 33: Thirty-third character is `a', increment the counter (counter = 16).\\Step 34: Thirty-fourth character is `b', no `a' found.\\Step 35: Thirty-fifth character is `b', no `a' found.\\Step 36: Thirty-sixth character is `a', increment the counter (counter = 17).\\Step 37: Thirty-seventh character is `b', no `a' found.\\Step 38: Thirty-eighth character is `b', no `a' found.\\Step 39: Thirty-ninth character is `a', increment the counter (counter = 18).\\Step 40: Fortieth character is `b', no `a' found.\\Step 41: Forty-first character is `b', no `a' found.\\Step 42: Forty-second character is `a', increment the counter (counter = 19).\\\\Result: \textcolor{red}{19}
       \end{minipage}\\[0.8em]
    \midrule[1.2pt]
    
    \rowcolor{gray!10} \multicolumn{1}{p{\linewidth}}{\textbf{Correct Answer:}} \\[0.4em]
     18\\[0.4em]
    
    \end{tabular}
    \end{tcolorbox}
    \caption{Case Study 6 Continued}
    \label{tab:casestudyqwenincorrectcont}
    
\end{table*}

\begin{table*}[htbp]
    \centering
    \small
    \begin{tcolorbox}[
        colback=white,
        boxrule=1.2pt,
        arc=8pt,
        outer arc=8pt,
        width=0.95\textwidth
    ]
    \begin{tabular}{p{0.93\textwidth}}
     \rowcolor{gray!10} \multicolumn{1}{p{\linewidth}}{\textbf{Model \& Experiment:}} \\[0.4em]
    Qwen Turbo, token type (d), length 30-40 \\[0.4em]
    \midrule[1.2pt]
    \rowcolor{gray!10} \multicolumn{1}{p{\linewidth}}{\textbf{Prompt:}} \\[0.4em]
    \begin{minipage}{0.93\textwidth}
Task: Count the number of occurrences of the substring `a' in the given string step by step. Do not use bold font in the response.\\

    Instructions:
    \begin{itemize}
        \item Iterate through the string and count each occurrence of the substring.
        \item Use a counter to keep track of how many times the substring appears.
        \item Output your step-by-step counting process.
        \item Conclude with: `Result: ' followed by the number. 
    \end{itemize}

    String: [`b', `a', `a', `a', `b', `a', `a', `a', `b', `a', `a', `a', `b', `a', `b', `b', `b', `a', `b', `a', `b', `b', `a', `a', `b', `a', `b', `b', `a', `b', `b', `a', `a', `b', `b']\\\\
    Response:
    \end{minipage} \\[0.8em]
    \midrule[1.2pt]
     \rowcolor{gray!10} \multicolumn{1}{p{\linewidth}}{\textbf{Model Response:}} \\[0.4em]
    \begin{minipage}{0.93\textwidth}
    To count the number of occurrences of the substring `a' in the provided string, we'll go through it step by step:\\\\- First element: `b' - No match. Counter = 0\\- Second element: `a' - Match found. Counter = 1\\- Third element: `a' - Match found. Counter = 2\\- Fourth element: `a' - Match found. Counter = 3\\- Fifth element: `b' - No match. Counter = 3\\- Sixth element: 'a' - Match found. Counter = 4\\- Seventh element: `a' - Match found. Counter = 5\\- Eighth element: `a' - Match found. Counter = 6\\- Ninth element: `b' - No match. Counter = 6\\- Tenth element: `a' - Match found. Counter = 7\\- Eleventh element: `a' - Match found. Counter = 8\\- Twelfth element: `a' - Match found. Counter = 9\\- Thirteenth element: `b' - No match. Counter = 9\\- Fourteenth element: `a' - Match found. Counter = 10\\- Fifteenth element: `b' - No match. Counter = 10\\- Sixteenth element: `b' - No match. Counter = 10\\- Seventeenth element: `b' - No match. Counter = 10\\- Eighteenth element: `a' - Match found. Counter = 11\\- Nineteenth element: `b' - No match. Counter = 11\\- Twentieth element: `a' - Match found. Counter = 12\\- Twenty-first element: `b' - No match. Counter = 12\\- Twenty-second element: `b' - No match. Counter = 12\\- Twenty-third element: `a' - Match found. Counter = 13\\- Twenty-fourth element: `a' - Match found. Counter = 14\\- Twenty-fifth element: `b' - No match. Counter = 14\\- Twenty-sixth element: `a' - Match found. Counter = 15\\- Twenty-seventh element: `b' - No match. Counter = 15\\- Twenty-eighth element: `b' - No match. Counter = 15\\- Twenty-ninth element: `a' - Match found. Counter = 16\\- Thirtieth element: `b' - No match. Counter = 16\\- Thirty-first element: `b' - No match. Counter = 16\\- Thirty-second element: `a' - Match found. Counter = 17\\- Thirty-third element: `a' - Match found. Counter = 18\\- Thirty-fourth element: `b' - No match. Counter = 18\\\\Result: \textcolor{teal}{18}
       \end{minipage}\\[0.8em]
    \midrule[1.2pt]
    
    \rowcolor{gray!10} \multicolumn{1}{p{\linewidth}}{\textbf{Correct Answer:}} \\[0.4em]
     18\\[0.4em]
    
    \end{tabular}
    \end{tcolorbox}
    \caption{Case Study 7 - Correct Prediction with Qwen Turbo (CoT prompt)}
    \label{tab:casestudyqwencorrect}
\end{table*}

\begin{table*}[htbp]
    \centering
    \small
    \begin{tcolorbox}[
        colback=white,
        boxrule=1.2pt,
        arc=8pt,
        outer arc=8pt,
        width=0.95\textwidth
    ]
    \begin{tabular}{p{0.93\textwidth}}
     \rowcolor{gray!10} \multicolumn{1}{p{\linewidth}}{\textbf{Model \& Experiment:}} \\[0.4em]
    GPT-4o mini, token type (a), length 20-30 \\[0.4em]
    \midrule[1.2pt]
    \rowcolor{gray!10} \multicolumn{1}{p{\linewidth}}{\textbf{Prompt:}} \\[0.4em]
    \begin{minipage}{0.93\textwidth}
Count the number of appearances of 'a's in the string below. Directly output 'Result: ' followed by the counted number. Do not use bold font in the response.\\

    String: abaaababbbbbbaaaabaaaa\\

    Response:
    \end{minipage} \\[0.8em]
    \midrule[1.2pt]
     \rowcolor{gray!10} \multicolumn{1}{p{\linewidth}}{\textbf{Model Response:}} \\[0.4em]
    \begin{minipage}{0.93\textwidth}
    Result: \textcolor{red}{10}
       \end{minipage}\\[0.8em]
    \midrule[1.2pt]
    
    \rowcolor{gray!10} \multicolumn{1}{p{\linewidth}}{\textbf{Correct Answer:}} \\[0.4em]
    13
     \\[0.4em]
    
    \end{tabular}
    \end{tcolorbox}
    \caption{Case Study 8 - Incorrect Prediction with GPT-4o mini (base prompt)}
    \label{tab:casestudygpt4ominiincorrect1}
\end{table*}

\clearpage
\begin{table*}[htbp]
    \centering
    \small
    \begin{tcolorbox}[
        colback=white,
        boxrule=1.2pt,
        arc=8pt,
        outer arc=8pt,
        width=0.95\textwidth
    ]
    \begin{tabular}{p{0.93\textwidth}}
     \rowcolor{gray!10} \multicolumn{1}{p{\linewidth}}{\textbf{Model \& Experiment:}} \\[0.4em]
    GPT-4o mini, token type (a), length 20-30 \\[0.4em]
    \midrule[1.2pt]
    \rowcolor{gray!10} \multicolumn{1}{p{\linewidth}}{\textbf{Prompt:}} \\[0.4em]
    \begin{minipage}{0.93\textwidth}
Task: Count the number of occurrences of the substring `a' in the given string step by step. Do not use bold font in the response.

    Instructions:
    \begin{itemize}
        \item Iterate through the string and count each occurrence of the substring.
        \item Use a counter to keep track of how many times the substring appears.
        \item Output your step-by-step counting process.
        \item Conclude with: `Result: ' followed by the number.
    \end{itemize}

    String: abaaababbbbbbaaaabaaaa
    \end{minipage} \\[0.8em]
    \midrule[1.2pt]
     \rowcolor{gray!10} \multicolumn{1}{p{\linewidth}}{\textbf{Model Response:}} \\[0.4em]
    \begin{minipage}{0.93\textwidth}
     1. Start with the string: abaaababbbbbbaaaabaaaa.\\2. Initialize a counter at 0.\\3. Iterate through each character in the string.\\\\   - Index 0: 'a' (counter = 1)\\   - Index 1: 'b' (counter remains 1)\\   - Index 2: 'a' (counter = 2)\\   - Index 3: 'a' (counter = 3)\\   - Index 4: 'a' (counter = 4)\\   - Index 5: 'b' (counter remains 4)\\   - Index 6: 'a' (counter = 5)\\   - Index 7: 'b' (counter remains 5)\\   - Index 8: 'b' (counter remains 5)\\   - Index 9: 'b' (counter remains 5)\\   - Index 10: 'b' (counter remains 5)\\   - Index 11: 'a' (counter = 6)\\   - Index 12: 'a' (counter = 7)\\   - Index 13: 'a' (counter = 8)\\   - Index 14: 'b' (counter remains 8)\\   - Index 15: 'a' (counter = 9)\\   - Index 16: 'a' (counter = 10)\\   - Index 17: 'a' (counter = 11)\\   - Index 18: 'a' (counter = 12)\\\\4. End of string reached.\\\\Result: \textcolor{red}{12}
       \end{minipage}\\[0.8em]
    \midrule[1.2pt]
    
    \rowcolor{gray!10} \multicolumn{1}{p{\linewidth}}{\textbf{Correct Answer:}} \\[0.4em]
    13
     \\[0.4em]
    
    \end{tabular}
    \end{tcolorbox}
    \caption{Case Study 9 - Incorrect Prediction with GPT 4o mini (CoT prompt)}
    \label{tab:casestudygpt4ominiincorrect2}
\end{table*}

\begin{table*}[htbp]
    \centering
    \small
    \begin{tcolorbox}[
        colback=white,
        boxrule=1.2pt,
        arc=8pt,
        outer arc=8pt,
        width=0.95\textwidth
    ]
    \begin{tabular}{p{0.93\textwidth}}
     \rowcolor{gray!10} \multicolumn{1}{p{\linewidth}}{\textbf{Model \& Experiment:}} \\[0.4em]
    GPT-4o mini, token type (a), length 20-30 \\[0.4em]
    \midrule[1.2pt]
    \rowcolor{gray!10} \multicolumn{1}{p{\linewidth}}{\textbf{Prompt:}} \\[0.4em]
    \begin{minipage}{0.93\textwidth}
Task: Count the number of occurrences of the substring `a' in the given string step by step. Do not use bold font in the response.\\

    Instructions:
    \begin{itemize}
        \item Iterate through the string and count each occurrence of the substring.
        \item Use a counter to keep track of how many times the substring appears.
        \item Output your step-by-step counting process.
        \item Conclude with: `Result: ' followed by the number.
    \end{itemize}

    String: [`a', `b', `a', `a', `a', `b', `a', `b', `b', `b', `b', `b', `b', `a', `a', `a', `a', `b', `a', `a', `a', `a']
    \end{minipage} \\[0.8em]
    \midrule[1.2pt]
     \rowcolor{gray!10} \multicolumn{1}{p{\linewidth}}{\textbf{Model Response:}} \\[0.4em]
    \begin{minipage}{0.93\textwidth}
    Let's count the occurrences of the substring `a' step by step in the given string:\\\\1. Initialize a counter at 0.\\2. Start iterating through the list: [`a', `b', `a', `a', `a', `b', `a', `b', `b', `b', `b', `b', `b', `a', `a', `a', `a', `b', `a', `a', `a', `a'].\\\\- Index 0: The element is `a'. Increment counter to 1.\\- Index 1: The element is `b'. Counter remains 1.\\- Index 2: The element is `a'. Increment counter to 2.\\- Index 3: The element is `a'. Increment counter to 3.\\- Index 4: The element is `a'. Increment counter to 4.\\- Index 5: The element is 'b'. Counter remains 4.\\- Index 6: The element is `a'. Increment counter to 5.\\- Index 7: The element is `b'. Counter remains 5.\\- Index 8: The element is `b'. Counter remains 5.\\- Index 9: The element is `b'. Counter remains 5.\\- Index 10: The element is `b'. Counter remains 5.\\- Index 11: The element is `b'. Counter remains 5.\\- Index 12: The element is `b'. Counter remains 5.\\- Index 13: The element is `a'. Increment counter to 6.\\- Index 14: The element is `a'. Increment counter to 7.\\- Index 15: The element is `a'. Increment counter to 8.\\- Index 16: The element is `a'. Increment counter to 9.\\- Index 17: The element is `b'. Counter remains 9.\\- Index 18: The element is `a'. Increment counter to 10.\\- Index 19: The element is `a'. Increment counter to 11.\\- Index 20: The element is `a'. Increment counter to 12.\\- Index 21: The element is `a'. Increment counter to 13.\\\\3. The final count after iterating through all elements is 13.\\\\Result: \textcolor{teal}{13}
       \end{minipage}\\[0.8em]
    \midrule[1.2pt]
    
    \rowcolor{gray!10} \multicolumn{1}{p{\linewidth}}{\textbf{Correct Answer:}} \\[0.4em]
    13
     \\[0.4em]
    
    \end{tabular}
    \end{tcolorbox}
    \caption{Case Study 10 - Correct Prediction with GPT-4o mini (CoT prompt)}
    \label{tab:casestudygpt4ominicorrect}
\end{table*}

\end{document}